\documentclass[11pt]{article}
\usepackage{acl2012}
\usepackage{times}
\usepackage{latexsym}
\usepackage{amsmath}
\usepackage{multirow}
\usepackage{url}
\usepackage{graphicx}
\usepackage[ruled,vlined]{algorithm2e}
\usepackage{amssymb}
\usepackage{amsfonts}
\usepackage{todonotes}
\usepackage{fancyvrb}
\usepackage{tikz}
\usetikzlibrary{bayesnet}
\usepackage[font=small,skip=3pt]{caption}

\makeatletter
\g@addto@macro{\small}{%
   \setlength{\abovedisplayskip}{0pt} 
   \setlength{\belowdisplayskip}{0pt} }
\makeatother

\makeatletter
\g@addto@macro{\scriptsize}{%
   \setlength{\abovedisplayskip}{0pt} 
   \setlength{\belowdisplayskip}{0pt} }
\makeatother

\makeatletter
\g@addto@macro{\footnotesize}{%
   \setlength{\abovedisplayskip}{0pt} 
   \setlength{\belowdisplayskip}{0pt} }
\makeatother

\makeatletter
\newcommand{\@BIBLABEL}{\@emptybiblabel}
\newcommand{\@emptybiblabel}[1]{}
\makeatother
\usepackage[hidelinks]{hyperref}

\newcommand{\citep}[1]{\cite{#1}}
\newcommand{\citet}[1]{\newcite{#1}}

\DeclareMathOperator*{\argmax}{arg\,max}
\DeclareMathAlphabet{\mathbfit}{OML}{cmm}{b}{it}

\usepackage{xcolor}
\newcommand{\CHANGEA}[1]{#1}   % this version does not show the changes

\newcommand{\CHANGEB}[1]{#1}   % this version does not show the changes

\newcommand{\CHANGEC}[1]{#1}  % this version shows the changes

\newcommand{\Ber}{\ensuremath{\mathrm{Ber}}}
\newcommand{\Cat}{\ensuremath{\mathrm{Cat}}}
\newcommand{\Dir}{\ensuremath{\mathrm{Dir}}}

\renewcommand{\Pr}{\ensuremath{\mathrm{P}}}

\newcommand{\Tr}[1]{\ensuremath{{#1}^{\scriptscriptstyle\top}}}
\newcommand{\SV}[1]{\ensuremath{\boldsymbol{#1}}}
\newcommand{\V}[1]{\ensuremath{\mathbfit{#1}}}

\newcommand{\CatE}{\ensuremath{\mathrm{CatE}}}

\newcommand{\Mstem}{M}
\newcommand{\MM}[2]{\ensuremath{{\Mstem^{#2}_{#1}}}}
\newcommand{\Nstem}{N}
\newcommand{\NN}[2]{\ensuremath{\Nstem^{#2}_{#1}}}
\newcommand{\Kstem}{K}
\newcommand{\KK}[2]{\ensuremath{\Kstem^{#2}_{#1}}}

\pdfpageattr{/Group << /S /Transparency /I true /CS /DeviceRGB>>}

\title{Improving Topic Models with Latent Feature Word Representations}

\author{Dat Quoc Nguyen${}^{1}$, Richard Billingsley${}^{1}$, Lan Du${}^{1}$ \and Mark Johnson${}^{1,2}$ \\
${}^{1}$ Department of Computing, Macquarie University, Sydney, Australia \\
${}^{2}$ Santa Fe Institute, Santa Fe, New Mexico, USA \\
{\tt{\footnotesize{dat.nguyen@students.mq.edu.au, \{richard.billingsley, lan.du, mark.johnson\}@mq.edu.au}}}
}

\begin{document}
\maketitle

\begin{abstract}
Probabilistic topic models are widely used to discover  latent topics in document collections, while latent feature vector representations of words have been used to obtain high performance in many NLP tasks.  In this paper, we extend two \CHANGEA{different} Dirichlet multinomial topic models by incorporating latent feature vector representations of words trained on very large corpora to improve the word-topic mapping learnt on a smaller corpus. Experimental results show that by using information from the external corpora, our new models produce significant  improvements on topic coherence, document clustering and document classification tasks, especially on datasets with few or short documents.
\end{abstract}

%\vspace{-10pt}

\section{Introduction}

Topic modeling algorithms, such as Latent Dirichlet Allocation  \cite{Blei2003} and related methods \cite{Blei2012}, are often used to learn a set of latent topics for a corpus, and predict the probabilities of each word in each document belonging to each topic \cite{TehNW2006,Newman:2006:SEM:1150402.1150487,NIPS2007_964,PorteousNIASW2008,johnson:2010:ACL,Pengtao13,Hingmire:2013:DCT:2484028.2484140}.

Conventional topic modeling algorithms such as these infer document-to-topic and topic-to-word
distributions from the co-occurrence of words within documents.  But when the training corpus of
documents is small or when the documents are short, the resulting distributions might be based on
little evidence.  \newcite{Sahami20006} and \newcite{Phan2011HTF} show that it  helps to exploit
external knowledge to improve the topic representations.  \newcite{Sahami20006} employed web search
results to improve the information in short texts. \newcite{Phan2011HTF} assumed that the small
corpus is a sample of topics from a larger corpus like Wikipedia, and then use the topics discovered
in the larger corpus to help shape the topic representations in the small corpus. However, if the
larger corpus has many irrelevant topics, this will \CHANGEA{``use up''} the topic space of the model.
In addition, \newcite{NIPS2010_4094} proposed an extension of LDA that uses external information
about word similarity, such as thesauri and dictionaries, to smooth the topic-to-word distribution.
 
Topic models have also been constructed using latent features
\cite{NIPS2009_3856,Nitish2013,AAAI159303}. Latent feature (\textsc{lf}) vectors have been used for
a wide range of NLP tasks \cite{ICML2011Glorot_342,socherEtAl2013,Pennington14}.  The combination of
values permitted by latent features forms a high dimensional space which makes it is well suited to
model topics of very large corpora.

Rather than relying solely on a multinomial or latent feature model, as in  \newcite{NIPS2009_3856},
\newcite{Nitish2013} and \newcite{AAAI159303}, we explore how to take advantage of both latent feature
and multinomial models by using a latent feature representation trained on a large external corpus
to supplement a multinomial topic model estimated from a smaller corpus.
% can be brought together by using the latent feature representation to
% incorporate external information and supplement the smaller corpus's multinomial topic representation.

Our main contribution is that we  propose two new latent feature topic models which integrate latent feature  word representations into two Dirichlet multinomial topic models: a Latent Dirichlet Allocation (\textsc{lda}) model \cite{Blei2003} and a one-topic-per-document Dirichlet Multinomial Mixture (\textsc{dmm}) model \cite{Nigam2000}. Specifically, we replace the topic-to-word Dirichlet multinomial component which generates the words from topics in each Dirichlet multinomial topic model by a two-component mixture of a Dirichlet multinomial component and a latent feature component. 

In addition to presenting a sampling procedure for the new models, we also compare using two different sets of pre-trained latent feature word vectors with our models.  We achieve significant improvements on topic coherence evaluation,  document clustering and document classification tasks, especially on corpora of short documents and corpora with few documents.

\section{Background}
\label{sec:background}

\subsection{\textsc{lda} model}

  The Latent Dirichlet Allocation (\textsc{lda}) topic model \cite{Blei2003} represents each document $d$ as a probability distribution $\SV\theta_d$ over topics, where each topic $z$ is modeled by a probability distribution $\SV\phi_z$ over words in a fixed vocabulary $W$.

\begin{figure}[ht]
\centering
$\begin{array}{cc}
\includegraphics[width=4cm]{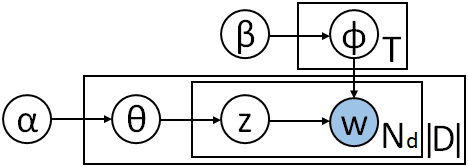} &
\includegraphics[width=3.5cm]{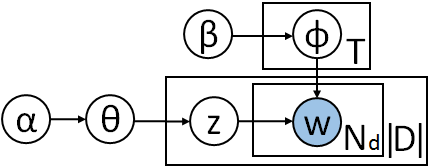} 
\\
\mbox{(\textsc{lda})} & \mbox{(\textsc{dmm})}
\end{array}$
\caption{Graphical models of \textsc{lda} and \textsc{dmm}}
\label{fig:graphicalmodels}
\end{figure}

  As presented in Figure \ref{fig:graphicalmodels}, where $\alpha$ and $\beta$ are hyper-parameters and $T$ is number of topics,   the generative process for \textsc{lda} is described as follows:

\medskip
{
\noindent
\begin{tabular}{p{4cm}l}
$\SV\theta_d \sim \Dir(\alpha) $ & $z_{d_i} \sim  \Cat(\SV\theta_d)$\\
$\SV\phi_z  \sim  \Dir(\beta)$ & $w_{d_i}  \sim \Cat(\SV\phi_{z_{d_i}})$\\
\end{tabular}
}
\medskip

\noindent where $\Dir$ and $\Cat$ stand for a Dirichlet distribution and a categorical distribution, and
$z_{d_i}$ is the topic indicator for the $i^{\mathit{th}}$ word $w_{d_i}$ in document $d$. Here, the topic-to-word Dirichlet multinomial component  generates the word $w_{d_i}$ by drawing it from the categorical distribution $\Cat(\SV\phi_{z_{d_i}})$  for topic $z_{d_i}$.

We follow the Gibbs sampling algorithm for estimating \textsc{lda} topic models as described by
\newcite{GriffithsS2004}. By integrating out $\SV\theta$ and $\SV\phi$, the algorithm samples the
topic $z_{d_i}$ for  the current $i^{\mathit{th}}$ word  $w_{d_i}$ in document $d$ using the conditional
distribution  $\Pr(z_{d_i} \mid \V{Z}_{\neg d_i})$, where $\V{Z}_{\neg d_i}$ denotes the topic
assignments of all the other words in the document collection $D$, so:

{\small
%\vspace{-15pt}
\begin{equation}\label{equation:Gibbs}
\begin{split}
 \Pr(z_{d_i} = t \mid \V{Z}_{\neg d_i}) \propto  (\NN{d_{\neg i}}{t} + \alpha) \frac{\NN{\neg d_i}{t,w_{d_i}} + \beta}{\NN{\neg d_i}{t} + V\beta} 
\end{split}
\end{equation}
%\vspace{-10pt}
}

\textbf{Notation}:
$\NN{d}{t,w}$ is the rank-3 tensor that counts the number of times that word $w$ is generated from topic $t$  in document $d$ by the Dirichlet multinomial component, which in  section 2.1 belongs to the \textsc{lda} model, while in section 2.2 belongs to the \textsc{dmm} model. When an index is omitted, it indicates summation over that index (so $N_d$ is the number of words in document $d$).

We write the subscript $\neg d$ for the document collection $D$ with document $d$ removed, and the subscript $\neg d_i$  for D with just the  $i^{\mathit{th}}$ word in  document $d$ removed, while the subscript ${d_{\neg i}}$ represents document $d$ without its $i^{\mathit{th}}$ word. For example, $N^t_{\neg d_i}$ is the number of words labelled a topic $t$, ignoring the $i^{\mathit{th}}$ word of document $d$.

$V$ is the size of the vocabulary, $V = |W|$.

\subsection{\textsc{dmm} model for short texts}

Applying topic models for short or few documents for text clustering is more challenging because of data sparsity and the limited contexts in such texts. One approach is to combine short texts into long pseudo-documents before training \textsc{lda} \cite{Hong2010D,Weng2010,Mehrotra2013}. Another approach is to assume that there is only one topic per document \cite{Nigam2000,Zhao2011,Yin2014}. 

In the Dirichlet Multinomial Mixture \textsc{(dmm)} model \cite{Nigam2000}, each document is assumed to only have one topic. The process of generating a document $d$ in the collection $D$, as shown in Figure \ref{fig:graphicalmodels}, is to first select a topic assignment for the document, and then the topic-to-word Dirichlet multinomial component generates all the words in the document from the same selected topic:

\medskip
{
\noindent
\begin{tabular}{p{4cm}l}
$\SV\theta \sim \Dir(\alpha) $ & $z_{d} \sim  \Cat(\SV\theta)$\\
$\SV\phi_z  \sim  \Dir(\beta)$ & $w_{d_i}  \sim \Cat(\SV\phi_{z_{d}})$\\
\end{tabular}
}
\medskip

\newcite{Yin2014} introduced a collapsed Gibbs sampling algorithm  for the \textsc{dmm} model in which a topic $z_d$ is sampled for the document $d$ using the conditional probability $\Pr(z_d \mid \textbf{Z}_{\neg d})$, where $\textbf{Z}_{\neg d}$ denotes the topic assignments of all the other documents, so:

{\scriptsize
%\vspace{-10pt}
\begin{equation}
\begin{split}
 & \Pr(z_d = t \mid \textbf{Z}_{\neg d}) \propto \\ 
  &  (M_{\neg d}^{t} + \alpha) \frac{\Gamma(\NN{\neg d}{t} + V\beta)}{ \Gamma(\NN{\neg d}{t} + \NN{d}{} + V\beta)} \prod_{w\in W}\frac{\Gamma(\NN{\neg d}{t,w} +\NN{d}{w} + \beta)}{ \Gamma(\NN{\neg d}{t,w} + \beta)}
\end{split}
\label{equa:GSdmm}
\end{equation}
}
%\vspace{-10pt}

\textbf{Notation:} $M_{\neg d}^{t}$ is the number of documents assigned to topic $t$ excluding the current document $d$; $\Gamma$ is the  Gamma function.% with $\Gamma(x+1) = x\Gamma(x)$.

\subsection{Latent feature vector models}

Traditional count-based methods \cite{deer1990lsa,Lund96,Bullinaria07} for learning real-valued latent feature (\textsc{lf}) vectors rely on co-occurrence counts. Recent approaches based on deep neural networks learn vectors by predicting words given their window-based context \cite{Collobert2008,MikolovNIPS2013,Pennington14,AAAI159314}.

 \newcite{MikolovNIPS2013}'s method maximizes the log likelihood of each word given its context. 
 \newcite{Pennington14} used back-propagation to minimize the squared error of a prediction of the log-frequency of context words within a fixed window of each word. 
 Word vectors can be trained directly on a new corpus. %, while pre-trained vectors trained on a large news corpus are also available. 
In  our new models, however, in order to incorporate the rich information from very large datasets,  we utilize pre-trained word vectors  that were trained on external billion-word corpora.

\section{New latent feature topic models}

In this section, we propose two novel probabilistic topic models, which we call the \textsc{lf-lda} and the \textsc{lf-dmm}, that combine a latent feature model with either an \textsc{lda} or \textsc{dmm} model. We also present Gibbs sampling procedures for our new models. 

\noindent
\begin{figure}[ht]
%\vspace{-10pt}
\centering
$\begin{array}{cc}
\includegraphics[width=3.75cm]{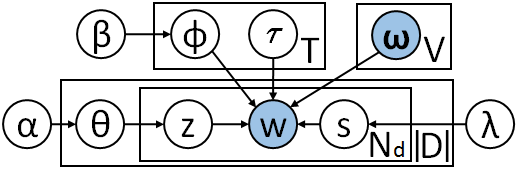} &
\includegraphics[width=3.75cm]{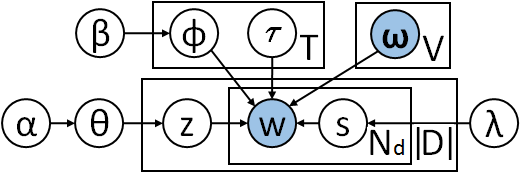} 
\\
\mbox{(\textsc{lf-lda})} & \mbox{(\textsc{lf-dmm})}
\end{array}$
\caption{Graphical models of our combined models}
\label{fig:ourlmodels}
\end{figure}

In general,  \textsc{lf-lda} and \textsc{lf-dmm} are formed by taking the original Dirichlet multinomial topic models \textsc{lda} and \textsc{dmm}, and replacing their topic-to-word Dirichlet multinomial component that generates words from topics with a two-component mixture of a topic-to-word Dirichlet multinomial component and a latent feature component. 

Informally, the new models have the structure of the original Dirichlet multinomial topic models, as shown in Figure \ref{fig:ourlmodels}, with the addition of two matrices $\SV\tau$ and $\SV\omega$ of latent feature weights, where $\SV\tau_t$ and $\SV\omega_w$ are the latent-feature vectors associated with
topic $t$ and word $w$ respectively.

Our latent feature model defines the probability that it generates a word given the topic as the categorical distribution $\CatE$ with: 

{\small
%\vspace{-15pt}
\begin{equation}
\CatE(w \mid \SV\tau_t\Tr{\SV\omega}) = \frac{\exp(\SV\tau_t \cdot \SV\omega_w )}{\sum_{w' \in W} \exp( \SV\tau_t \cdot \SV\omega_{w'})}
\end{equation}
%\vspace{-10pt}
}

$\CatE$ is a categorical distribution with log-space parameters, i.e. { $\CatE(w \mid  \V{u})\propto\exp({u}_w)$}.
 As $\SV\tau_t$ and $\SV\omega_w$ are (row) vectors of latent feature
weights, so $\SV\tau_t
\Tr{\SV\omega}$ is a vector of ``scores'' indexed by words. %, so $\Pr_{\textit{{\Tiny \textsc{lf}}}}(w|z=t,\SV\omega,\SV\tau)=\CatE(w|\SV\tau_t\Tr{\SV\omega})$.
  $\SV\omega$ is fixed because we use pre-trained word vectors. 
  
In the next two sections \ref{ssec:lflda} and \ref{ssec:lfdmm}, we explain the generative processes of our new models  \textsc{lf-lda} and  \textsc{lf-dmm}.  
  We then present our Gibbs sampling procedures for the models \textsc{lf-lda} and \textsc{lf-dmm} in the  sections \ref{gibbslflda} and \ref{gibbslfdmm}, respectively, and explain how we estimate $\SV\tau$ in section \ref{subsec:topicvector}.

\subsection{Generative process for the \textsc{lf-lda} model}
\label{ssec:lflda}

The \textsc{lf-lda} model generates a document as follows: a distribution over topics $\SV\theta_d$ is drawn for  document $d$; then for each $i^{\mathit{th}}$ word $w_{d_i}$ (in sequential order that words appear in the document), the model chooses a topic indicator $z_{d_i}$, 
 a binary indicator variable $s_{d_i}$ is sampled from a Bernoulli distribution to determine whether the word $w_{d_i}$ is to be generated by the Dirichlet multinomial or latent feature component, and finally  the word is generated from the chosen topic by the determined topic-to-word model. The generative process is:
 
\medskip
\noindent
\begin{tabular}{p{4cm}l}
$\SV\theta_d \sim \Dir(\alpha) $ & $z_{d_i} \sim  \Cat(\SV\theta_d)$\\
$\SV\phi_z  \sim  \Dir(\beta)$ & $s_{d_i}  \sim  \Ber(\lambda)$ \\
\multicolumn{2}{l}{{\small $w_{d_i}  \sim (1-s_{d_i}) \Cat(\SV\phi_{z_{d_i}}) + s_{d_i} \CatE(\SV\tau_{z_{d_i}}\,\Tr{\SV\omega})$}}
\end{tabular}

\medskip

\noindent where the hyper-parameter $\lambda$ is the probability of a word being generated by the latent feature topic-to-word model and $\Ber(\lambda)$ is a Bernoulli distribution with success probability $\lambda$.

\subsection{Generative process for the \textsc{lf-dmm} model}
\label{ssec:lfdmm}

Our \textsc{lf-dmm} model  uses the \textsc{dmm} model assumption that all the words in a document share the same topic. Thus, the process of generating a document in a document collection with our \textsc{lf-dmm} is as follows: a distribution over topics $\SV\theta$ is drawn for the document collection; then the model draws a topic indicator $z_{d}$ for the entire document $d$; 
 for every $i^{\mathit{th}}$ word $w_{d_i}$ in the document $d$, a binary indicator variable $s_{d_i}$ is sampled from a Bernoulli distribution to determine whether the Dirichlet multinomial or latent feature component will be used to generate the word $w_{d_i}$, and finally the word is generated from the same topic $z_{d}$ by the determined component. The generative process is summarized as:

\medskip
\noindent
\begin{tabular}{p{4cm}l}
$\SV\theta \sim \Dir(\alpha) $ & $z_d \sim  \Cat(\SV\theta)$\\
$\SV\phi_z  \sim  \Dir(\beta)$ & $s_{d_i}  \sim  \Ber(\lambda)$ \\
\multicolumn{2}{l}{$w_{d_i}  \sim (1-s_{d_i}) \Cat(\SV\phi_{z_d}) + s_{d_i} \CatE(\SV\tau_{z_d}\,\Tr{\SV\omega}) $}
\end{tabular}

\subsection{Inference in \textsc{lf-lda}  model}
\label{gibbslflda}

From the generative model of \textsc{lf-lda} in Figure \ref{fig:ourlmodels}, by integrating out $\SV\theta$ and $\SV\phi $, we use the Gibbs sampling algorithm  \cite{Robert2004} to perform inference to calculate the conditional topic assignment probabilities for each word. The outline of the Gibbs sampling algorithm for the \textsc{lf-lda} model is detailed in Algorithm \ref{alg:Gibbs_lflda}.  

\begin{algorithm}[!ht]
\SetAlgoVlined
{\small
Initialize the word-topic variables $z_{d_i}$ using the \textsc{lda} sampling algorithm

\For(){iteration iter = 1, 2, ...}{

\For(){topic t = 1, 2, ..., T}{$\SV\tau_t = \argmax_{\SV\tau_t} \Pr(\SV\tau_t \mid \V{Z}, \V{S})$}

\For(){document d = 1, 2, ..., $|D|$}{

\For(){word index i = 1, 2, ..., $N_d$}{

sample $z_{d_i}$ and $s_{d_i}$ from $\Pr(z_{d_i} = t, s_{d_i} \mid \V{Z}_{\neg d_i}, \V{S}_{\neg d_i}, \SV\tau, \SV\omega)$
}
}
}
}
\caption{An approximate Gibbs sampling algorithm for the \textsc{lf-lda} model}
\label{alg:Gibbs_lflda}
\end{algorithm}

%\vspace{-10pt}

Here, $\V{S}$ denotes the distribution indicator variables for the whole document collection $D$.  Instead of sampling $\SV\tau_t$ from the posterior, we perform MAP estimation as \CHANGEA{described} in the section \ref{subsec:topicvector}.

For sampling the topic $z_{d_i}$ and the binary indicator variable $s_{d_i}$ of the $i^{\mathit{th}}$ word $w_{d_i}$ in the document $d$, we integrate out $s_{d_{i}}$  in order to sample $z_{d_{i}}$  and then sample $s_{d_{i}}$ given $z_{d_{i}}$.  We sample the topic $z_{d_i}$ using the conditional distribution as follows:

%\vspace{-5pt}
{\small
\begin{equation}
\label{equation:Gibbs}
\begin{split}
 & \Pr(z_{d_i} = t \mid \V{Z}_{\neg d_i}, \SV\tau, \SV\omega)  \\ 
 & \propto  (\NN{d _{\neg i}}{t} + \KK{d_{\neg i}}{t} + \alpha) 
 \\ & \left(
  (1-\lambda) \frac{\NN{\neg d_i}{t,w_{d_i}}+\beta}{\NN{\neg d_i}{t}+V\beta}
  + \lambda \CatE(w_{d_i} \mid \SV\tau_{t}\,\Tr{\SV\omega}) 
\right)
\end{split}
\end{equation}
%\vspace{-10pt}
}

Then we sample ${s}_{d_i}$ conditional on $z_{d_i}=t$ with:

{\small
\begin{equation}
\resizebox{.89\hsize}{!}{$
\Pr(s_{d_i}{=}s \mid z_{d_i}{=}t) \propto \left\{ 
  \begin{array}{l}
  (1-\lambda) \frac{\NN{\neg d_i}{t,w_{d_i}}+\beta}{\NN{\neg d_i}{t}+V\beta}\;\text{for $s$ = 0}\\
 \lambda\;\CatE(w_{d_i}|\SV\tau_{t}\,\Tr{\SV\omega})\;\text{for $s$ = 1}
  \end{array} \right.
  $}
\end{equation}
}
 
\textbf{{Notation}}: 
Due to the new models' mixture architecture, we separate out the counts for each of the two components of each model. We define the rank-3 tensor $\KK{d}{t,w}$ as the number of times a word $w$ in document $d$ is generated from topic $t$ by the latent feature component of the generative \textsc{lf-lda}  or  \textsc{lf-dmm} model. 

We also extend the earlier definition of the tensor $\NN{d}{t,w}$ as the number of times a word $w$ in document $d$ is generated from topic $t$ by the Dirichlet multinomial component of our combined models, which in section \ref{gibbslflda} refers to the \textsc{lf-lda} model, while in section \ref{gibbslfdmm} refers to the \textsc{lf-dmm} model.  For both tensors $K$ and $N$, omitting an index refers to summation over that index and negation $\neg$ indicates exclusion as before. So $\NN{d}{w} + \KK{d}{w}$ is the total number of times the word type $w$ appears in the document $d$.

\subsection{Inference in \textsc{lf-dmm}  model}
\label{gibbslfdmm}

 For the \textsc{lf-dmm} model, we  integrate out $\SV\theta$ and $\SV\phi$, and then sample the topic $z_d$ and the distribution selection variables $\V{s}_d$ for document $d$ using Gibbs sampling as outlined in Algorithm \ref{alg:Gibbs_lfdmm}.
 
 %\vspace{-5pt}

\begin{algorithm}[!ht]
\SetAlgoVlined
{\small
Initialize the word-topic variables $z_{d_i}$ using the \textsc{dmm} sampling algorithm

\For(){iteration iter = 1, 2, ...}{

\For(){topic t = 1, 2, ..., T}{$\SV\tau_t = \argmax_{\SV\tau_t} \Pr(\SV\tau_t \mid \V{Z}, \V{S})$}

\For(){document d = 1, 2, ..., $|D|$}{
sample $z_d$ and  $\V{s}_d$ from $\Pr(z_d=t, \V{s}_d \mid \V{Z}_{\neg d}, \V{S}_{\neg d}, \SV\tau, \SV\omega)$
}
}
}
\caption{An approximate Gibbs sampling algorithm for the \textsc{lf-dmm} model}
\label{alg:Gibbs_lfdmm}
\end{algorithm}

As before in Algorithm \ref{alg:Gibbs_lflda}, we also use MAP estimation of $\SV\tau$ as detailed in section \ref{subsec:topicvector} rather than sampling from the posterior. The conditional distribution of topic variable and selection variables for document $d$ is:
%

%\vspace{-10pt}

{\small
\begin{equation}
\resizebox{.875\hsize}{!}{$
\begin{split}
& \Pr(z_d=t, \V{s}_d \mid \V{Z}_{\neg d},  \V{S}_{\neg d}, \SV\tau, \SV\omega) \\
 & \propto  \lambda^{\KK{d}{}}\;(1-\lambda)^{\NN{d}{}}\;
 (\MM{\neg d}{t}+\alpha)  \frac{\Gamma(\NN{\neg d}{t} + V \beta)}{\Gamma(\NN{\neg d}{t} + \NN{d}{} + V\beta)} \\
 & \ \ \  \ \prod_{w \in W} \frac{\Gamma(\NN{\neg d}{t,w} + \NN{d}{w} + \beta)}{\Gamma(\NN{\neg d}{t,w} + \beta)}  \prod_{w \in W} \CatE(w \mid \SV\tau_{t}\,\Tr{\SV\omega})^{\KK{d}{w}}
\end{split}
 $}
\end{equation}
}

%\vspace{-5pt}

Unfortunately the ratios of Gamma functions makes
it difficult to integrate out $\V{s}_d$ in
this distribution $\Pr$. As $z_d$ and $\V{s}_d$ are not independent, it is
computationally expensive to directly sample  from \CHANGEA{this distribution, as there are} $2^{(\NN{d}{w} + \KK{d}{w})}$
different values of $\V{s}_d$.  So we approximate $\Pr$ with a distribution $Q$ that factorizes
across words as follows:

%\vspace{-3pt}

{\footnotesize
\setlength{\abovedisplayskip}{0pt}
\begin{eqnarray}
\lefteqn{Q(z_d=t, \V{s}_d \mid \V{Z}_{\neg d}, \V{S}_{\neg d}, \SV\tau, \SV\omega)} \nonumber \\
 & \propto & \lambda^{\KK{d}{}}\;(1-\lambda)^{\NN{d}{}}\;
 (\MM{\neg d}{t}+\alpha ) \\ 
 & & \prod_{w \in W} \left(\frac{\NN{\neg d}{t,w}+\beta}{\NN{\neg d}{t}+V\beta}\right)^{\NN{d}{w}}  \prod_{w \in W} \CatE(w \mid \SV\tau_{t}\,\Tr{\SV\omega})^{\KK{d}{w}} \nonumber
\end{eqnarray}
}
%\vspace{-3pt}

%\textbf{REASON to sample from Q in practical implementation}

This simpler distribution $Q$ can be viewed as an approximation to $\Pr$ in which the topic-word
``counts" are ``frozen" within a document. This approximation is reasonably accurate for short
documents. \CHANGEA{\CHANGEB{This distribution $Q$ simplifies the coupling between $z_d$ and $\V{s}_d$}. This enables us to integrate out
$\V{s}_d$ in $Q$. We first sample the document topic $z_d$ for
document $d$ using  $Q(z_d)$, marginalizing over $s_d$}:
%

%\vspace{-5pt}

{\footnotesize
\begin{eqnarray}
  \lefteqn{Q(z_d = t\mid \V{Z}_{\neg d}, \SV\tau, \SV\omega)} \nonumber \\
  & \propto & (\MM{\neg d}{t}+\alpha)  \prod_{w\in W} \left(
  \begin{array}{l}
  (1-\lambda) \frac{\NN{\neg d}{t,w}+\beta}{\NN{\neg d}{t}+V\beta} \\
  + \;\lambda\;\CatE(w \mid \SV\tau_{t}\,\Tr{\SV\omega}) 
  \end{array}\right)^{(\NN{d}{w}+\KK{d}{w})}
\end{eqnarray}
}

%\vspace{-5pt}

Then we sample the binary indicator variable $s_{d_i}$ for each $i^{\mathit{th}}$ word $w_{d_i}$ in document $d$ conditional on $z_d=t$ \CHANGEA{from the following distribution}:

{\small
\begin{equation}
\resizebox{.89\hsize}{!}{$
Q(s_{d_i}{=}s \mid z_d = t) \propto \left\{ 
  \begin{array}{l}
  (1-\lambda) \frac{\NN{\neg d}{t,w_{d_i}}+\beta}{\NN{\neg d}{t}+V\beta}\;\text{for $s$ = 0}\\
 \lambda\;\CatE(w_{d_i} \mid \SV\tau_{t}\,\Tr{\SV\omega})\;\text{for $s$ = 1}
  \end{array} \right.
  $}
\end{equation}
}

%\vspace{-5pt}

\subsection{Learning latent feature vectors for topics}
\label{subsec:topicvector}

To estimate the topic vectors after each Gibbs sampling iteration through the data, \CHANGEB{we apply regularized maximum likelihood estimation}. Applying MAP estimation to learn log-linear models for \CHANGEB{topic models} \CHANGEA{is also used } in  SAGE \cite{ICML2011Eisenstein_534} and SPRITE \cite{TACL403}. However, unlike our
models, those models do not use latent feature word vectors \CHANGEA{to characterize topic-word distributions}.  The negative log
likelihood of the corpus $L$ under our model  factorizes topic-wise into factors $L_t$ for each topic. With
$L_2$ regularization\footnote{The $L_2$ regularizer constant \CHANGEA{was set to} $\mu = 0.01$.} for topic $t$, \CHANGEA{these are}: 

%\vspace{-5pt}
{\footnotesize
\begin{equation}
\resizebox{.89\hsize}{!}{$
\begin{split}
L_t = - \sum_{w \in W} & \KK{}{t,w} \Big(\SV\tau_t \cdot \SV\omega_w - \log \big( \sum_{w' \in W} \exp(\SV\tau_t \cdot \SV\omega_{w'}) \big) \Big) \\ 
& + \mu\parallel\SV\tau_t\parallel^2_2
\end{split}
$}
\end{equation}
}

The MAP estimate of topic vectors $\SV\tau_t$ is obtained by minimizing the regularized negative log likelihood. The derivative with respect to the $j^{\mathit{th}}$ element of the vector for topic $t$ is:

\vspace{-2pt}

{\footnotesize
%\vspace{-5pt}
\begin{equation}
\resizebox{.89\hsize}{!}{$
\begin{split}
\frac{\partial L_t}{\partial \SV\tau_{t,j}} = - \sum_{w \in W} & \KK{}{t,w} \Big( \SV\omega_{w,j} - \sum_{w' \in W} \SV\omega_{w',j} \CatE(w' \mid \SV\tau_t\Tr{\SV\omega}) \Big) \\
& + 2 \mu  \SV\tau_{t,j}
\end{split}
$}
\end{equation}
%\vspace{-5pt}
}

\vspace{-2pt}

We used \textsc{l-bfgs}\footnote{We used the L-BFGS implementation from the Mallet toolkit \cite{AndrewMcCallum2002}.}% We set the L-BFGS convergence condition as  $|L_t^{l+1} - L_t^{l}| \leq 0.05$ where the index $l$ denotes the negative log likelihood value $L_t$ at the $l^{th}$ iteration.} 
 \cite{Liu1989} to find the topic vector $\SV\tau_t$ that \CHANGEA{minimizes} $L_t$.

\section{Experiments}

To investigate the performance of our new  \textsc{lf-lda} and \textsc{lf-dmm} models, we compared their performance against baseline  \textsc{lda} and \textsc{dmm} models on topic coherence, document clustering and document classification \CHANGEA{evaluations}. The topic coherence evaluation measures the coherence of topic-word associations, i.e. it directly evaluates how coherent  the assignment of words to topics is. \CHANGEA{The document clustering and document classification tasks evaluate how useful the topics  assigned to documents are in clustering and classification tasks}. 

Because we expect our new models to perform comparatively well in situations where there is little data about topic-to-word distributions, \CHANGEA{our experiments focus on corpora with few or short documents}. We also investigated which values of $\lambda$ perform well, and compared the performance when using two different sets of pre-trained word vectors in these new models.
 
\subsection{Experimental setup}

\subsubsection{Distributed word representations}

% \newcite{Marco14} showed that deep learning-based distributed representation models can outperform the traditional count-based word vector models on lexical semantics tasks. Therefore, we experimented with two state-of-the-art sets of deep learning-based pre-trained word vectors here.

\noindent We experimented with two state-of-the-art sets of pre-trained word vectors here.
  
Google word vectors\footnote{{\scriptsize Download at: \url{https://code.google.com/p/word2vec/}}}  are pre-trained 300-dimensional vectors for 3 million words and phrases. These vectors were trained on a 100 billion word subset of  the Google News corpus by using the Google {Word2Vec} toolkit \cite{MikolovNIPS2013}. 
 Stanford vectors\footnote{{\scriptsize Download at: \url{http://www-nlp.stanford.edu/projects/glove/}}}  are pre-trained 300-dimensional  vectors for 2 million  words. These vectors were learned \CHANGEA{from} 42-billion tokens of Common Crawl  web data \CHANGEA{using} the Stanford {GloVe} toolkit \cite{Pennington14}.

We refer to our \textsc{lf-lda} and \textsc{lf-dmm}  models  using Google and Stanford word vectors as \textbf{w2v-\textsc{lda}}, \textbf{glove-\textsc{lda}}, \textbf{w2v-\textsc{dmm}} and \textbf{glove-\textsc{dmm}}. 

\subsubsection{Experimental datasets}

\noindent We conducted experiments on the 20-Newsgroups dataset, the TagMyNews news dataset and the Sanders Twitter corpus.

 The 20-Newsgroups  dataset\footnote{We used the ``all-terms'' version of the 20-Newsgroups dataset available at \url{http://web.ist.utl.pt/acardoso/datasets/} \cite{Ana-Cardoso-Cachopo}.} contains about 19,000 newsgroup  documents evenly grouped into 20 different categories. The TagMyNews news dataset\footnote{The TagMyNews news dataset is  unbalanced, where the largest  category contains 8,200 news items while the smallest category contains about 1,800 items. Download at: \url{http://acube.di.unipi.it/tmn-dataset/}} \cite{Vitale2012} consists of about 32,600 English RSS news items grouped into 7 categories, where each news document has a news title and a short description. 
In our experiments, \CHANGEA{we also used a news title dataset which consists of just the news titles from the  TagMyNews news dataset}. 

Each dataset was down-cased, and we removed non-alphabetic characters and stop-words found in the
stop-word list in the Mallet toolkit  \cite{AndrewMcCallum2002}. We also removed words shorter than
3 characters and words appearing less than 10 times in the 20-Newsgroups \CHANGEA{corpus}, and under 5 times in the
TagMyNews news and news titles datasets. In addition, words not found in both Google and Stanford
vector representations were \CHANGEA{also} removed.\footnote{1366, 27 and 12 words were correspondingly removed out of the 20-Newsgroups, TagMyNews news and news title datasets.} 
We refer to the cleaned 20-Newsgroups, TagMyNews news and news title datasets as \textbf{N20}, \textbf{TMN} and \textbf{TMNtitle}, respectively.
  
We also performed experiments on two subsets of the N20 dataset. 
The \textbf{N20short} dataset consists of all documents from the N20 dataset \CHANGEA{with less than 21 words}. 
The  \textbf{N20small} dataset contains 400 documents consisting of 20 randomly selected documents from each group \CHANGEA{of} the N20 dataset.

\begin{table}[h]
%\vspace{-5pt}
\centering
{\footnotesize
\begin{tabular}{lllll}
\hline
Dataset  &\#g & \#docs & \#w/d & $V$\\
\hline
N20  & 20 & 18,820 & 103.3 & 19,572 \\
N20short  & 20 & 1,794 & 13.6 & 6,377 \\
N20small  & 20 & 400 & 88.0 & 8,157 \\
TMN  & 7  & 32,597 & 18.3 & 13,428\\
TMNtitle  & 7 & 32,503 & 4.9 & 6,347 \\
Twitter  & 4 & 2,520 & 5.0 & 1,390\\
\hline
\end{tabular}
}
\caption{Details of experimental datasets. \#g: number of ground truth labels; \#docs: number of documents; \#w/d: the average number of words per document; $V$: the number of word types}
%\vspace{-5pt}
\label{tab:datasets}
\end{table}

Finally, we also experimented  on the  publicly available Sanders Twitter corpus.\footnote{Download
  at: \url{http://www.sananalytics.com/lab/index.php}} This corpus consists of 5,512 Tweets grouped into
four different topics (Apple, Google, Microsoft, and Twitter). Due to restrictions in Twitter's
Terms of Service, the actual Tweets need to be downloaded using 5,512 Tweet IDs. There are 850
Tweets not available to download. After removing the non-English Tweets, 3,115 Tweets remain. In
addition to converting into lowercase and removing non-alphabetic characters, words were normalized
by using a lexical normalization dictionary for microblogs \cite{han2012}. We then removed
stop-words, words shorter than 3 characters or appearing less than 3 times in the corpus. The four
words  \textit{apple}, \textit{google}, \textit{microsoft} and \textit{twitter} were removed as
these four words occur in every Tweet in the corresponding topic. Moreover, words not found in both
Google and Stanford \CHANGEA{vector lists} were also removed.\footnote{There are 91 removed words.} In all our experiments, after removing words from documents, \CHANGEA{any document with a zero word count was also removed from the corpus}. For the Twitter corpus, this resulted in just 2,520 remaining Tweets.

\subsubsection{General settings}

\noindent The hyper-parameter  $\beta$ used in baseline \textsc{lda} and \textsc{dmm} models was set to 0.01, as this is a common setting in the literature \cite{GriffithsS2004}. \CHANGEA{ We set the hyper-parameter  $\alpha = 0.1$, as this can improve performance relative to the standard setting $\alpha=\frac{50}{T}$, as noted by \newcite{LuMZ2011} and \newcite{Yin2014}.}

We ran each baseline model for 2000  iterations and evaluated the topics assigned to words in the last sample. \CHANGEA{For our models, we ran the baseline models for 1500 iterations, then used the outputs from the last sample to initialize our models, which we ran for 500 further iterations}. %\footnote{We found that it took 1500 iterations for the \textsc{lda} model to reach convergence on the N20 dataset. Initialized by the \textsc{lda} model, our \textsc{lf-lda} model took 500 further iterations to converge on  the N20 dataset. On other datasets, all 4 experimental models required a smaller number of iterations to reach convergence. So, we set all models to run for a total of 2000 iterations. }. 

We report the mean and standard deviation of the results of ten repetitions of each experiment (\CHANGEA{so  the standard deviation is approximately 3 standard errors, or a 99\% confidence interval}).%\footnote{For the baseline models, the experimental results obtained for the outputs from the $2000^{th}$ sample are similar to the ones from the $1500^{th}$ sample.}.

\subsection{Topic coherence evaluation}
This section examines the quality of the topic-word mappings induced by our models. In our models, topics are distributions over words. The topic coherence evaluation measures to what extent the high-probability words in each topic are semantically coherent \cite{NIPS2009_3700,Stevens2012}. 

\subsubsection{Quantitative analysis}

\noindent \newcite{Newman2010}, \newcite{Mimno2011} and \newcite{Lau2014} \CHANGEA{describe} methods for automatically evaluating the semantic coherence of sets of words. \CHANGEA{The method presented in \newcite{Lau2014} uses the normalized pointwise mutual information \textsc{(npmi)} score}  and has a strong correlation with human-judged coherence. A higher \textsc{npmi} score indicates that the topic distributions are semantically more coherent. Given a topic $t$ represented by its top-$N$ topic words $w_1, w_2, ..., w_{N}$, the \textsc{npmi} score for  $t$ is:
 
\vspace{-5pt}
 
{\small
\begin{equation}
\label{equal:npmi}
\begin{split}
\text{NPMI-Score}(t) = \sum_{1 \leqslant i< j \leqslant N} \frac{\log \frac{\Pr(w_i, w_j)}{\Pr(w_i)\Pr(w_j)}}{-\log \Pr(w_i, w_j)}
\end{split}
\end{equation}
}

\noindent where the probabilities in equation (\ref{equal:npmi}) are derived \CHANGEA{from a 10-word sliding window over an external corpus}.

The \textsc{npmi} score for a topic model is  the average  score for all topics. We compute the \textsc{npmi} score based on top-15 most probable words of each topic and use the English Wikipedia\footnote{We used the Wikipedia-articles dump \CHANGEA{of} July 8, 2014.} of  4.6 million articles as \CHANGEA{our} external corpus.

 \begin{figure}[!ht]
\centering
\includegraphics[width=8cm,height=4.5cm]{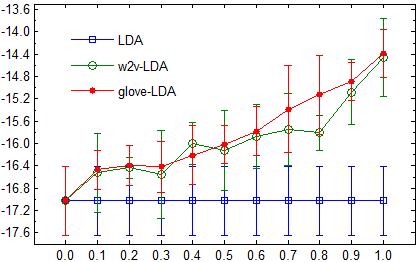}
\caption{ \textsc{npmi} scores (mean and standard deviation) on the N20short dataset with 20 topics, varying the mixture weight $\lambda$ from 0.0 to 1.0.}
\label{fig:NPMIN20short20T}
\end{figure}

%\vspace{-15pt}

 \begin{figure}[!ht]
\centering
\includegraphics[width=8cm,height=4.5cm]{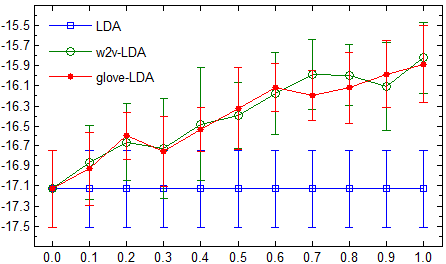}
\caption{\textsc{npmi} scores  on the N20short dataset with 40 topics, varying the mixture weight $\lambda$ from 0.0 to 1.0.}
\label{fig:NPMIN20short40T}
\end{figure}

%\vspace{-5pt}

Figures \ref{fig:NPMIN20short20T} and \ref{fig:NPMIN20short40T} show \textsc{npmi} scores computed for the \textsc{lda}, w2v-\textsc{lda} and glove-\textsc{lda} models on the N20short dataset for 20 and 40 topics. \CHANGEA{We} see that $\lambda =1.0$ gives the highest \textsc{npmi} score. In other words, using only the latent feature model produces the most coherent topic distributions. %However, the combined model \textsc{lf-lda} with $\lambda =0.6$ still returns a significant improvement compared to the baseline model\footnote{The improvement is significant according to a student's t-Test with a p-value less than 0.01.}.

\begin{table}[!ht]
\centering
%{\scriptsize
%\def\arraystretch{1.05}
\setlength{\tabcolsep}{0.25em}
\resizebox{8.125cm}{!}{
\begin{tabular}{l|l|l|l|l|l}
\hline
\multirow{2}{*}{{ Data}} & \multirow{2}{*}{Method}&  \multicolumn{3}{|c}{$\lambda = 1.0$}\\
\cline{3-6}
&  & T=6 & T=20 &  T=40 & T=80 \\
\hline
\hline
 & \textsc{lda} & -16.7 $\pm$ 0.9 & -11.7 $\pm$ 0.7 & -11.5 $\pm$ 0.3 & -11.4 $\pm$ 0.4\\
N20 & w2v-\textsc{lda} & -14.5 $\pm$ 1.2 & -9.0 $\pm$ 0.8 & -10.0 $\pm$ 0.5 & -10.7 $\pm$ 0.4 \\
 & glove-\textsc{lda} & \textbf{-11.6} $\pm$ 0.8 & \textbf{-7.4} $\pm$ 1.0 & \textbf{-8.3} $\pm$ 0.7 & \textbf{-9.7} $\pm$ 0.4\\
\cline{2-6}
& Improve.  & 5.1 & 4.3 & 3.2 & 1.7\\
\hline
\hline
 & \textsc{lda} & -18.4 $\pm$ 0.6 & -16.7 $\pm$ 0.6 & -17.8 $\pm$ 0.4 & -17.6 $\pm$ 0.3\\
N20small & w2v-\textsc{lda} & \textbf{-12.0} $\pm$ 1.1 &  \textbf{-12.7}  $\pm$ 0.7 & -15.5  $\pm$ 0.4 & \textbf{-16.3}  $\pm$ 0.3\\
 & glove-\textsc{lda} & -13.0 $\pm$ 1.1 & -12.8  $\pm$ 0.7 & \textbf{-15.0}  $\pm$ 0.5 & -16.6  $\pm$ 0.2\\
\cline{2-6}
& Improve. & 6.4 & 4.0 & 2.8 & 1.3 \\
\hline
\end{tabular}
%}
}
\caption{\textsc{npmi} scores (mean and standard deviation) for N20 and N20small datasets. The
  \textit{Improve.} row denotes the absolute  improvement accounted for the best result produced by our latent feature model over the baselines.}
\label{tab:npmi}
\end{table}

\begin{table}[!ht]
\centering
%{\scriptsize
%\def\arraystretch{1.05}
\setlength{\tabcolsep}{0.2em}
\resizebox{8.125cm}{!}{
\begin{tabular}{l|l|l|l|l|l}
\hline
\multirow{2}{*}{{ Data}} & \multirow{2}{*}{Method}&  \multicolumn{4}{|c}{$\lambda = 1.0$}\\
\cline{3-6}
&  & T=7 & T=20 &  T=40 & T=80 \\
\hline
\hline
 & \textsc{lda}& -17.3 $\pm$ 1.1 & -12.7 $\pm$ 0.8 & -12.3 $\pm$ 0.5 & -13.0 $\pm$ 0.3 \\
TMN & w2v-\textsc{lda} & -14.7 $\pm$ 1.5 & -12.8 $\pm$ 0.8 & -12.2 $\pm$ 0.5 & -13.1 $\pm$ 0.2\\
& glove-\textsc{lda} & \textbf{-13.0} $\pm$ 1.8 & \textbf{-9.7} $\pm$ 0.7 & \textbf{-11.5} $\pm$ 0.5 & \textbf{-12.9} $\pm$ 0.4\\
\cline{2-6}
& Improve. & 4.3 & 3.0 & 0.8 & 0.1\\
\hline
 & \textsc{dmm} & -17.4 $\pm$ 1.5 & -12.2 $\pm$ 1.0 & -10.6 $\pm$ 0.6 & -11.2 $\pm$ 0.4 \\
TMN & w2v-\textsc{dmm} & \textbf{-11.5} $\pm$ 1.6 & -7.0 $\pm$ 0.7 & \textbf{-5.8} $\pm$ 0.5 & \textbf{-5.8} $\pm$ 0.3\\
 & glove-\textsc{dmm} & -13.4 $\pm$ 1.5 & \textbf{-6.2} $\pm$ 1.2 & -6.6 $\pm$ 0.5 & -6.3 $\pm$ 0.5 \\
\cline{2-6}
& Improve. & 5.9 & 6.0 & 4.8 & 5.4 \\
\hline
\hline
 & \textsc{lda} &  -17.2 $\pm$ 0.8 & -15.4 $\pm$ 0.7 & -15.3 $\pm$ 0.3 & -15.6 $\pm$ 0.3 \\
TMNtitle & w2v-\textsc{lda} &  -14.2 $\pm$ 1.0 & -14.0 $\pm$ 0.7 & \textbf{-15.0} $\pm$ 0.3 & \textbf{-14.9} $\pm$ 0.4  \\
& glove-\textsc{lda} &  \textbf{-13.9} $\pm$ 0.9 & \textbf{-13.4} $\pm$ 0.7 & -15.2 $\pm$ 0.5 & -15.2 $\pm$ 0.2\\
\cline{2-6}
& Improve. & 3.3 & 2.0 & 0.3 & 0.7 \\
\hline
 & \textsc{dmm} &  -16.5 $\pm$ 0.9 & -13.6 $\pm$ 1.0 & -13.1 $\pm$ 0.5 & -13.7 $\pm$ 0.3\\
TMNtitle & w2v-\textsc{dmm} & \textbf{-9.6} $\pm$ 0.6 & \textbf{-7.5} $\pm$ 0.8 & \textbf{-8.1} $\pm$ 0.4 & -9.7 $\pm$ 0.4\\
& glove-\textsc{dmm}  &  -10.9 $\pm$ 1.3 & -8.1 $\pm$ 0.5 & \textbf{-8.1} $\pm$ 0.5 & \textbf{-9.1} $\pm$ 0.3\\
\cline{2-6}
& Improve. & 5.6 & 6.1 & 5.0 & 4.6\\
\hline 
\end{tabular}
%}
}
\caption{\textsc{npmi} scores for TMN  and TMNtitle datasets.}% \textit{Improve.} denotes the absolute  improvement accounted for the best result (in bold) produced by our latent feature model over the baseline models.}
\label{tab:npmi2}
\end{table}

\begin{table}[!ht]
\centering
%{\scriptsize
%\def\arraystretch{1.05}
\setlength{\tabcolsep}{0.2em}
\resizebox{8.125cm}{!}{
\begin{tabular}{l|l|l|l|l|l}
\hline
\multirow{2}{*}{{ Data}} & \multirow{2}{*}{Method}&  \multicolumn{4}{|c}{$\lambda = 1.0$}\\
\cline{3-6}
&  & T=4 & T=20 &  T=40 & T=80 \\
\hline
\hline
 & \textsc{lda} & -8.5 $\pm$ 1.1 & -14.5 $\pm$ 0.4 & -15.1 $\pm$ 0.4 & -15.9 $\pm$ 0.2\\
Twitter & w2v-\textsc{lda} & -7.3 $\pm$ 1.0 & \textbf{-13.2} $\pm$ 0.6 & \textbf{-14.0} $\pm$ 0.3 & \textbf{-14.1} $\pm$ 0.3 \\
& glove-\textsc{lda} & \textbf{-6.2} $\pm$ 1.6 & -13.9 $\pm$ 0.6 & -14.2 $\pm$ 0.4 & -14.2 $\pm$ 0.2 \\
\cline{2-6}
& Improve.  & 2.3 & 1.3 & 1.1 & 1.8\\
\hline
 & \textsc{dmm} & -5.9 $\pm$ 1.1 & -10.4 $\pm$ 0.7 & -12.0 $\pm$ 0.3 & -13.3 $\pm$ 0.3\\
Twitter & w2v-\textsc{dmm} & -5.5 $\pm$ 0.7 & -10.5 $\pm$ 0.5 & -11.2 $\pm$ 0.5 & \textbf{-12.5} $\pm$ 0.1\\
 & glove-\textsc{dmm} & \textbf{-5.1} $\pm$ 1.2 & \textbf{-9.9} $\pm$ 0.6 & \textbf{-11.1}  $\pm$ 0.3 & \textbf{-12.5} $\pm$ 0.4\\
\cline{2-6}
& Improve.  & 0.8 & 0.5 & 0.9 & 0.8 \\
\hline
\end{tabular}
%}
}
\caption{\textsc{npmi} scores for Twitter  dataset.}% \textit{Improve.} denotes the absolute  improvement accounted for the best result (in bold) produced by our latent feature model over the baseline models.}
\label{tab:npmi3}
\end{table}

Tables \ref{tab:npmi}, \ref{tab:npmi2} and \ref{tab:npmi3} present the \textsc{npmi} scores produced by the models on the \CHANGEA{other} experimental datasets, where we vary\footnote{ We perform with $T=6$ on the N20 and N20small datasets as the 20-Newsgroups dataset could be also grouped into 6 larger topics instead of 20 fine-grained categories.} the number of topics in steps from $4$ to $80$. \CHANGEA{Tables} \ref{tab:npmi2} and \ref{tab:npmi3} show that the \textsc{dmm} model performs better than the \textsc{lda} model on the \CHANGEA{TMN,  TMNtitle and Twitter datasets. These results show} that our latent feature models produce significantly higher scores than the baseline models on all the experimental datasets. 

%\textbf{Influence of $\lambda$ and $T$:} For smaller values of $T$, for example 7 or 20,  setting $\lambda$ to 1.0 noticeably improves the topic-word mapping compared with  $\lambda=0.6$ . However, for larger values of $T$ , the \textsc{npmi} scores become less sensitive to variations of  $\lambda$ between 0.6 and 1.0. Particularly, on the small dataset N20small with $T=80$, $\lambda=0.6$  returns a better \textsc{npmi} score than $\lambda=1.0$.

\textbf{Google word2vec vs. Stanford glove word vectors:} In general, our latent feature models obtain competitive \textsc{npmi} results in using pre-trained Google word2vec and Stanford glove word vectors for a large value of $T$, for example $T=80$. 
 With small values of $T$, for example  $T \leq 7$ , using Google word vectors produces better
 scores than using Stanford word vectors on the small N20small dataset of normal texts and on the
 short text TMN and TMNtitle  datasets. However, the opposite pattern holds on the full N20 dataset.
\CHANGEA{Both sets of the pre-trained word vectors produce similar scores on the
 small and short Twitter dataset}.

\begin{table*}[!ht]
\centering
%{\scriptsize
%\def\arraystretch{1.025}
\setlength{\tabcolsep}{0.35em}
\resizebox{16.5cm}{!}{
\begin{tabular}{llllllllllll}
\hline
\multicolumn{9}{c|}{Topic 1} & \multicolumn{3}{c}{Topic 3} \\
\hline
Init\textsc{dmm} & Iter=1 & Iter=2  & Iter=5 & Iter=10 &  Iter=20 & Iter=50 & Iter=100 & {Iter=500} & \multicolumn{1}{|l}{Init\textsc{dmm}} & Iter=50 & Iter=500 \\
\hline

{japan} & japan  & japan   &  japan  & japan & japan  & japan &  japan & {japan}  & \multicolumn{1}{|l}{{u.s.}}  & prices & {prices} \\
{nuclear} &  nuclear  & nuclear & nuclear &  nuclear & nuclear  & nuclear &  nuclear &  {nuclear} &  \multicolumn{1}{|l}{{oil}} & sales & {sales} \\
{u.s.} &  u.s.  & u.s.  & u.s. &  u.s.  & u.s.  & plant &  u.s.  & {u.s.}  & \multicolumn{1}{|l}{{japan}} & oil &  {oil} \\
{crisis}  & \textbf{russia}  & crisis   &  plant &  plant  &  plant &  u.s.  & plant &  {plant}   & \multicolumn{1}{|l}{{prices}} & u.s. & {u.s.} \\
{plant}  & radiation & china   & crisis &  radiation  &  \textbf{quake}  & \textbf{quake} &  \textbf{quake}  & \textbf{quake} & \multicolumn{1}{|l}{{stocks}}  & stocks & {profit} \\
\underline{china}  & \textbf{nuke} &  \textbf{russia} &  radiation &  crisis & radiation  & radiation &  radiation  & {radiation}  &  \multicolumn{1}{|l}{{sales}}  & profit & {stocks} \\
\underline{libya}  & \textbf{iran} &  plant    & china  & china & crisis  & \textbf{earthquake} & \textbf{earthquake} & \textbf{earthquake}  &  \multicolumn{1}{|l}{{profit}}  & japan & {japan}\\
{radiation}  & crisis  & radiation  &  \textbf{russia} &  \textbf{nuke}  &  \textbf{nuke}  & \textbf{tsunami} &  \textbf{tsunami}   & \textbf{tsunami}  &  \multicolumn{1}{|l}{\underline{fed}}  & rise & {rise} \\
\underline{u.n.}  & china  & \textbf{nuke}  & \textbf{nuke}  & \textbf{russia} & china  & \textbf{nuke}  &  \textbf{nuke}  & \textbf{nuke}   &  \multicolumn{1}{|l}{{rise}} &  \textbf{gas} & \textbf{gas} \\
\underline{vote} &  libya  & libya   & \textbf{power}  & \textbf{power}  & \textbf{tsunami}  & crisis  & crisis  & {crisis}     & \multicolumn{1}{|l}{{growth}}  & growth & {growth} \\
\underline{korea} &  plant  & \textbf{iran}    & u.n.  & u.n. & \textbf{earthquake}  & \textbf{disaster} &  \textbf{disaster} & \textbf{disaster} &  \multicolumn{1}{|l}{\underline{wall}} & \textbf{profits}  & {shares} \\
\underline{europe} &  u.n.  & u.n.   & \textbf{iran}  & \textbf{iran} &  \textbf{disaster}  & \textbf{plants} &  \textbf{oil}  &   \textbf{power}  &  \multicolumn{1}{|l}{\underline{street}} & shares &  \textbf{price}\\
\underline{government} &  \textbf{mideast} &  \textbf{power}   & \textbf{reactor} & \textbf{earthquake} & \textbf{power} &  \textbf{power}  & \textbf{plants}  &  \textbf{oil}  &  \multicolumn{1}{|l}{\underline{china}} &\textbf{price} & \textbf{profits} \\
\underline{election}  & \textbf{pakistan} &  \textbf{pakistan}  & \textbf{earthquake} &  \textbf{reactor}  &  \textbf{reactor} &  \textbf{oil} &  \textbf{power}  &  \textbf{japanese}  & \multicolumn{1}{|l}{\underline{fall}} & \textbf{rises} &  \textbf{rises} \\
\underline{deal}  & \textbf{talks}  &  \textbf{talks}  &  libya  &  \textbf{quake} & \textbf{japanese}  & \textbf{japanese} &  \textbf{tepco}  & \textbf{plants}  & \multicolumn{1}{|l}{{shares}}  & \textbf{earnings} &  \textbf{earnings} \\ 
%\hline
%-22.67 & -14.66 & -16.80  & -14.46 & -13.54 & -8.80 & -11.16 & -6.18 & -11.16 & -11.16 & -22.41 & -15.14\\
\hline 
\hline
  \multicolumn{3}{c|}{Topic 4} &  \multicolumn{3}{c|}{Topic 5}  & \multicolumn{3}{c|}{Topic 19} & \multicolumn{3}{c}{Topic 14} \\
\hline
{Init\textsc{dmm}} &  Iter=50 & {Iter=500} & \multicolumn{1}{|l}{Init\textsc{dmm}} &  Iter=50 & {Iter=500} & \multicolumn{1}{|l}{Init\textsc{dmm}} &  Iter=50 & {Iter=500} & \multicolumn{1}{|l}{Init\textsc{dmm}} &  Iter=50 & {Iter=500} \\
\hline                                       
 {egypt} & libya  & {libya} & \multicolumn{1}{|l}{\underline{critic}} &   \textbf{dies}  &  {star}    &     \multicolumn{1}{|l}{{nfl}} & nfl  & {nfl} & \multicolumn{1}{|l}{\underline{nfl}}&  law  &  {law}  \\
  
\underline{china} & egypt  & {egypt} & \multicolumn{1}{|l}{\underline{corner}}  &  star  &  {sheen}   &        \multicolumn{1}{|l}{\underline{idol}}  & draft  & \textbf{sports} & \multicolumn{1}{|l}{\underline{court}}  & bill  &  {texas}   \\

\underline{u.s.} & \textbf{mideast}  & \textbf{iran}  & \multicolumn{1}{|l}{\underline{office}}  &  \textbf{broadway}  &  \textbf{idol}    &     \multicolumn{1}{|l}{{draft}} & \textbf{lockout}  & {draft} & \multicolumn{1}{|l}{{law}}  &  governor & {bill}  \\

{mubarak} & \textbf{iran} &  \textbf{mideast} & \multicolumn{1}{|l}{\underline{video}}  &  \textbf{american}  &  \textbf{broadway}   &     \multicolumn{1}{|l}{\underline{american}} &  players  & {players} & \multicolumn{1}{|l}{{bill}} & texas & {governor} \\

\underline{ bin} & \textbf{opposition} &  \textbf{opposition} & \multicolumn{1}{|l}{\underline{game}}  &  \textbf{idol} &   {show}     &     \multicolumn{1}{|l}{\underline{show}} &  \textbf{coach} & \textbf{lockout} & \multicolumn{1}{|l}{{wisconsin}} &  senate  &  {senate}   \\

 {libya} & \textbf{leader} &  \textbf{protests}  & \multicolumn{1}{|l}{{star}}  &  lady &   \textbf{american}   &       \multicolumn{1}{|l}{\underline{film}} &  \textbf{nba} & \textbf{football} & \multicolumn{1}{|l}{\underline{players}} & union  & {union}   \\
 
 \underline{laden} &  u.n.  & \textbf{leader}  & \multicolumn{1}{|l}{{lady}}  &  gaga &   {gaga}    &        \multicolumn{1}{|l}{\underline{season}} & \textbf{player}  & \textbf{league} & \multicolumn{1}{|l}{\underline{judge}} &  \textbf{obama} &   \textbf{obama}   \\
 
 \underline{france} & \textbf{protests} &  \textbf{syria} & \multicolumn{1}{|l}{{gaga}}  &  show  &  \textbf{tour}    &      \multicolumn{1}{|l}{\underline{sheen}} &  sheen &  {n.f.l.} & \multicolumn{1}{|l}{{governor}}  & wisconsin &  {budget}   \\
 
{bahrain} & \textbf{syria} &  {u.n.} & \multicolumn{1}{|l}{{show}}  &  \textbf{news}  &  \textbf{cbs}     &        \multicolumn{1}{|l}{{n.f.l.}} &  \textbf{league} & \textbf{player} & \multicolumn{1}{|l}{{union}} &  budget & {wisconsin}   \\

 \underline{air}  & \textbf{tunisia} &  \textbf{tunisia}  & \multicolumn{1}{|l}{\underline{weekend}}  &  critic  &  \textbf{hollywood}   &       \multicolumn{1}{|l}{\underline{back}} &  n.f.l. & \textbf{baseball} & \multicolumn{1}{|l}{\underline{house}} &  \textbf{state}  & \textbf{immigration}   \\ 
 
 \underline{report} & \textbf{protesters} &  \textbf{chief} & \multicolumn{1}{|l}{{sheen}}  &  \textbf{film} &   \textbf{mtv}     &      \multicolumn{1}{|l}{\underline{top}} &  \textbf{coaches}  & \textbf{court} & \multicolumn{1}{|l}{{texas}} &  \textbf{immigration}  &  \textbf{state}    \\
 
 \underline{rights}  & \textbf{chief} &  \textbf{protesters} & \multicolumn{1}{|l}{\underline{box}}  &  \textbf{hollywood}  &  {lady}     &            \multicolumn{1}{|l}{\underline{star}} &  \textbf{football} & \textbf{coaches} &  \multicolumn{1}{|l}{\underline{lockout}} &  \textbf{arizona}  & \textbf{vote}   \\
 
\underline{court} & \textbf{asia}  &  {mubarak} & \multicolumn{1}{|l}{\underline{park}}  &  \textbf{fame}  &  \textbf{wins}     &      \multicolumn{1}{|l}{\underline{charlie}} & \textbf{judge} & \textbf{nflpa} & \multicolumn{1}{|l}{{budget}} &  \textbf{california}  &  \textbf{washington}   \\

 {u.n.}  & \textbf{russia} &  \textbf{crackdown}  & \multicolumn{1}{|l}{\underline{takes}} &   \textbf{actor}  &  \textbf{charlie}    &      \multicolumn{1}{|l}{{players}} &  \textbf{nflpa} &  \textbf{basketball} & \multicolumn{1}{|l}{\underline{peru}} & \textbf{vote} &  \textbf{arizona}  \\
 
 \underline{war} & \textbf{arab} &  {bahrain} & \multicolumn{1}{|l}{\underline{man}}   & \textbf{movie}  &  \textbf{stars}     &       \multicolumn{1}{|l}{\underline{men}} & \textbf{court} &  \textbf{game} & \multicolumn{1}{|l}{{senate}} & \textbf{federal}  & \textbf{california}   \\
%\hline
% -18.04 & -5.84 & -0.45 & -18.37    &   -12.17   &   -10.82  &   -14.56 & -0.17 & 2.74 & -17.17 & -9.98 & -10.62  \\
\hline
\end{tabular}
%}
}
\caption{Examples of the 15 most probable topical words on the TMNtitle dataset with $T=20$.
  Init\textsc{dmm} denotes the output from the $1500^{\mathit{th}}$ sample produced by the
  \textsc{dmm} model, which we use to initialize the w2v-\textsc{dmm} model. Iter=$1$, Iter=$2$,
  Iter=$3$ and the like refer to the output of our w2v-\textsc{dmm} model after running $1$, $2$,
  $3$ sampling iterations, respectively. The words found in Init\textsc{dmm} and not found in
  Iter=$500$ are \underline{underlined}. \CHANGEA{ Words found by the w2v-\textsc{dmm} model but} not found by the \textsc{dmm} model are in \textbf{bold}. }
\label{tab:quanlity2}
\end{table*}

\subsubsection{Qualitative analysis}

\noindent \CHANGEA{This section provides an example of how our models improve topic coherence. Table \ref{tab:quanlity2}   compares the  top-15 words\footnote{In the baseline model, the top-15  topical words  output from the $1500^{\mathit{th}}$ sample are similar to top-15 words  from the $2000^{\mathit{th}}$ sample if we do not take the order of the most probable words into account.} produced by the baseline  \textsc{dmm} model  and our w2v-\textsc{dmm} model with $\lambda = 1.0$ on the TMNtitle dataset with $T = 20$ topics}.% It is clear that  our w2v-\textsc{dmm} model  forms  more coherent concepts than the \textsc{dmm} model. 

In table \ref{tab:quanlity2}, \CHANGEA{topic $1$ of the \textsc{dmm} model consists of words related to ``nuclear crisis in Japan'' together with other unrelated words. The w2v-\textsc{dmm} model produced a purer topic $1$ focused on ``Japan earthquake and nuclear crisis,'' presumably related to the ``Fukushima Daiichi nuclear disaster.''} %\footnote{\scriptsize{http://en.wikipedia.org/wiki/Fukushima\_Daiichi\_nuclear\_disaster}}. 
 Topic $3$ \CHANGEA{is about ``oil prices'' in} both models. However, all top-15 words are qualitatively more coherent in \CHANGEA{the w2v-\textsc{dmm}} model. 
While topic $4$ \CHANGEA{of} the \textsc{dmm} model is difficult to manually label, \CHANGEA{topic $4$ of the}  w2v-\textsc{dmm} model \CHANGEA{is about} the ``Arab Spring'' event.
 %\footnote{http://en.wikipedia.org/wiki/Arab\_Spring}. 

Topics $5$, $19$ and $14$ of the \textsc{dmm} model are not easy to label. Topic $5$ relates to ``entertainment'',  topic $19$ is generally a mixture of  ``entertainment'' and ``sport'', and topic $14$ is about ``sport'' and ``politics.''  
However, \CHANGEA{the  w2v-\textsc{dmm} model more clearly distinguishes these topics:  topic $5$  is  } about
``entertainment'', topic $19$ is only about ``sport'' and topic $14$ %\footnote{The topic $14$ reflects the ``2011 Wisconsin protests'' event: http://en.wikipedia.org/wiki/2011\_Wisconsin\_protests} 
 is only about ``politics.''

\subsection{Document clustering evaluation}

We compared our  models to the baseline models in a document clustering task. %This task is to examine the effectiveness of the most likely topic to clustering performance.
 After using a topic model to calculate the topic probabilities of a document, we assign every document  the topic with the highest probability given the document \cite{Cai2008,LuMZ2011,Pengtao13,Yan2013}.  
 We use two common metrics to evaluate  clustering performance: \textit{Purity} and \textit{normalized mutual information} \textsc{(nmi)}: \CHANGEA{see  \cite[Section 16.3]{ManningRS2008} for details} of these evaluations. Purity and \textsc{nmi} scores always range from 0.0 to 1.0, and higher scores reflect better clustering performance.
 
  \begin{figure}[!ht]
\centering
\includegraphics[width=8cm,height=4cm]{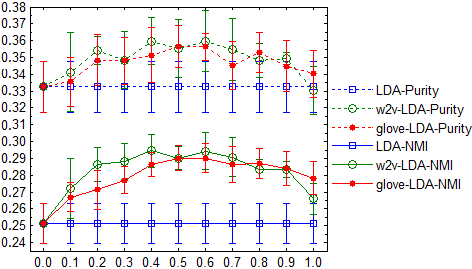}
\caption{Purity and \textsc{nmi} results (mean and standard deviation) on the N20short dataset with number of topics $T$ = 20, varying the mixture weight $\lambda$ from 0.0 to 1.0.}
\label{fig:N20short20T}
\end{figure}

 \begin{figure}[!ht]
\centering
\includegraphics[width=8cm,height=4cm]{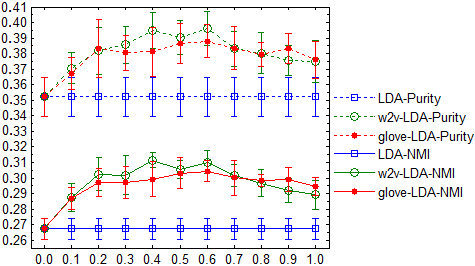}
\caption{Purity and \textsc{nmi} results on the N20short dataset with number of topics $T=40$, varying the mixture weight $\lambda$ from 0.0 to 1.0.}
\label{fig:N20short40T}
\end{figure}
 
\begin{table*}[ht]
\centering
%{\scriptsize
\resizebox{16.5cm}{!}{
\def\arraystretch{0.975}
\setlength{\tabcolsep}{0.2em}
\begin{tabular}{l|l|l|l|l|l|l|l|l|l}
\hline
\multirow{2}{*}{{ Data}} & \multirow{2}{*}{Method}& \multicolumn{4}{|c}{Purity} & \multicolumn{4}{|c}{NMI}\\
\cline{3-10}
 & & T=6 & T=20 &  T=40 & T=80 & T=6 & T=20 &  T=40 & T=80 \\
\hline
\hline
 & \textsc{lda} &  0.293 $\pm$ 0.002 & 0.573 $\pm$ 0.019  & \textbf{0.639} $\pm$ 0.017 & \textbf{0.646} $\pm$ 0.005 &  0.516 $\pm$ 0.009 & {0.582} $\pm$ 0.009 & \textbf{0.557} $\pm$ 0.007 & \textbf{0.515} $\pm$ 0.003\\
 
N20 & w2v-\textsc{lda} & 0.291 $\pm$ 0.002  & 0.569 $\pm$ 0.021 & 0.616 $\pm$ 0.017 & {0.638} $\pm$ 0.006 & 0.500 $\pm$ 0.008 & 0.563 $\pm$ 0.009 & 0.535 $\pm$ 0.008 & 0.505 $\pm$ 0.004\\

 & glove-\textsc{lda} & \textbf{0.295} $\pm$ 0.001 & \textbf{0.604} $\pm$ 0.031  & {0.632} $\pm$ 0.017 & {0.638} $\pm$ 0.007 & \textbf{0.522} $\pm$ 0.003  & \textbf{0.596} $\pm$ 0.012 & {0.550} $\pm$ 0.010 & {0.507} $\pm$ 0.003\\
\cline{2-10}
& Improve. &  0.002 & 0.031 & -0.007 & -0.008 &  0.006 & 0.014 & -0.007 & -0.008\\
\hline
\hline
  & \textsc{lda} & 0.232 $\pm$ 0.011 & 0.408 $\pm$ 0.017 & 0.477 $\pm$ 0.015 & 0.559 $\pm$ 0.018 &  0.376 $\pm$ 0.016 & 0.474 $\pm$ 0.013 & 0.513 $\pm$ 0.009 & 0.563 $\pm$  0.008\\   
N20small & w2v-\textsc{lda} & 0.229 $\pm$ 0.005 & \textbf{0.439} $\pm$ 0.015 & \textbf{0.516} $\pm$ 0.024 & \textbf{0.595} $\pm$ 0.016 & 0.406 $\pm$ 0.023 & \textbf{0.519} $\pm$ 0.014 & \textbf{0.548} $\pm$ 0.017 & \textbf{0.585} $\pm$ 0.009\\
    & glove-\textsc{lda} & \textbf{0.235} $\pm$ 0.008 & {0.427} $\pm$ 0.022 & {0.492} $\pm$ 0.022 & {0.579} $\pm$ 0.011 &  \textbf{0.436} $\pm$ 0.019& {0.504} $\pm$ 0.020 & {0.527} $\pm$ 0.013 & {0.576} $\pm$ 0.006\\
\cline{2-10}
& Improve. & 0.003 & 0.031 & 0.039 & 0.036 & 0.06 & 0.045 & 0.035 & 0.022 \\
\hline
\end{tabular}
%}
}
\caption{Purity and \textsc{nmi} results (mean and standard deviation) on the N20 and N20small datasets with  $\lambda = 0.6$. \textit{Improve.} row denotes the difference between the best result obtained by our model and the baseline model.}
%\vspace{-10pt}
\label{tab:puritynmi}
\end{table*}

\begin{table*}[!ht]
\centering
%{\scriptsize
\resizebox{16.5cm}{!}{
\def\arraystretch{0.975}
\setlength{\tabcolsep}{0.2em}
\begin{tabular}{l|l|l|l|l|l|l|l|l|l}
\hline
\multirow{2}{*}{{ Data}} & \multirow{2}{*}{Method}& \multicolumn{4}{|c}{Purity} & \multicolumn{4}{|c}{NMI}\\
\cline{3-10}
 & & T=7 & T=20 &  T=40 & T=80 & T=7 & T=20 &  T=40 & T=80 \\
\hline
\hline
 & \textsc{lda} & 0.648 $\pm$ 0.029 & 0.717 $\pm$ 0.009  & \textbf{0.721} $\pm$ 0.003  & 0.719 $\pm$ 0.007      & 0.436 $\pm$ 0.019 & 0.393 $\pm$ 0.008 & 0.354 $\pm$ 0.003 & 0.320 $\pm$ 0.003\\
 
TMN & w2v-\textsc{lda} & \textbf{0.658} $\pm$ 0.020 & 0.716 $\pm$ 0.012  & 0.720 $\pm$ 0.008  &  \textbf{0.725} $\pm$ 0.004 &   0.446 $\pm$ 0.014  & 0.399 $\pm$ 0.006  & 0.355 $\pm$ 0.005 & \textbf{0.325} $\pm$ 0.003 \\

 & glove-\textsc{lda} & \textbf{0.658} $\pm$ 0.034 & \textbf{0.722} $\pm$ 0.007  & 0.719 $\pm$ 0.008  &  \textbf{0.725} $\pm$ 0.006 &  \textbf{0.448} $\pm$ 0.017  & \textbf{0.403} $\pm$ 0.004 & \textbf{0.356} $\pm$ 0.004 & 0.324 $\pm$ 0.004 \\
\cline{2-10}
& Improve. & 0.01 & 0.005 & -0.001 & 0.006 & 0.012 & 0.01 & 0.002 & 0.005\\
\hline
%  & \textsc{dmm} & 0.632 $\pm$ 0.025 & 0.719 $\pm$ 0.020 & 0.735 $\pm$ 0.010 & 0.742 $\pm$ 0.005     & \textbf{0.445} $\pm$ 0.017 & 0.426 $\pm$ 0.010  & 0.397 $\pm$ 0.006 & 0.364 $\pm$ 0.002\\   
%TMN & w2v-\textsc{dmm} & 0.639 $\pm$ 0.024 & 0.741 $\pm$ 0.011 & 0.759 $\pm$ 0.006 & \textbf{0.778} $\pm$ 0.005 & 0.437 $\pm$ 0.018 & 0.429 $\pm$ 0.004 & 0.402 $\pm$ 0.003 & 0.377 $\pm$ 0.002  \\
%  & glove-\textsc{dmm} & \textbf{0.646} $\pm$ 0.022 & \textbf{0.757} $\pm$ 0.009 & \textbf{0.763} $\pm$ 0.005 & 0.775 $\pm$ 0.011 & \textbf{0.445} $\pm$ 0.023 & \textbf{0.443} $\pm$ 0.008 & \textbf{0.404} $\pm$ 0.003 & \textbf{0.378} $\pm$ 0.004\\
%\cline{2-10}
%& Improve. & 0.014 & 0.038 & 0.028 & 0.036 & 0.0 & 0.017 & 0.007 & 0.014 \\

  & \textsc{dmm} & 0.637 $\pm$ 0.029 & 0.699 $\pm$ 0.015 & 0.707 $\pm$ 0.014 & 0.715 $\pm$ 0.009 & 0.445 $\pm$ 0.024 & 0.422 $\pm$ 0.007 & 0.393 $\pm$ 0.009 & 0.364 $\pm$ 0.006\\   
TMN & w2v-\textsc{dmm} & 0.623 $\pm$ 0.020 & 0.737  $\pm$  0.018 &  \textbf{0.760}  $\pm$  0.010 & 0.772  $\pm$ 0.005 &  0.426  $\pm$  0.015 & 0.428  $\pm$  0.009 & 0.405  $\pm$  0.006 & 0.378  $\pm$  0.003 \\
  & glove-\textsc{dmm} & \textbf{0.641}  $\pm$  0.042  & \textbf{0.749}  $\pm$  0.011 & 0.758  $\pm$  0.008 & \textbf{0.776} $\pm$  0.006 & \textbf{0.449}  $\pm$  0.028 & \textbf{0.441}  $\pm$  0.008 & \textbf{0.408}  $\pm$  0.005 & \textbf{0.381}  $\pm$  0.003 \\
\cline{2-10}
& Improve. & 0.004 & 0.05  & 0.053 & 0.061 & 0.004 & 0.019 & 0.015 & 0.017   \\

\hline
\hline
 & \textsc{lda} & 0.572 $\pm$ 0.014 & 0.599 $\pm$ 0.015 & 0.593 $\pm$ 0.011 & 0.580 $\pm$ 0.006 & 0.314 $\pm$ 0.008 & 0.262 $\pm$ 0.006 & 0.228 $\pm$ 0.006 & 0.196 $\pm$ 0.003\\
TMNtitle & w2v-\textsc{lda} & {0.579} $\pm$ 0.020 & {0.619} $\pm$ 0.015 & \textbf{0.611} $\pm$ 0.007 & {0.598} $\pm$ 0.004 & {0.321} $\pm$ 0.012 & {0.279} $\pm$ 0.006 & \textbf{0.239} $\pm$ 0.005 & \textbf{0.210} $\pm$ 0.002\\
 & glove-\textsc{lda} & \textbf{0.584} $\pm$ 0.026 & \textbf{0.623} $\pm$ 0.012 & {0.600} $\pm$ 0.008 & \textbf{0.601} $\pm$ 0.004 & \textbf{0.322} $\pm$ 0.015 & \textbf{0.280} $\pm$ 0.004 & {0.235} $\pm$ 0.006 & {0.209} $\pm$ 0.003\\
\cline{2-10}
& Improve. & 0.012 & 0.024 & 0.018 & 0.021 & 0.008 & 0.018 & 0.011 & 0.014 \\
\hline 
% & \textsc{dmm} & 0.598 $\pm$ 0.018 & 0.650 $\pm$ 0.011 & 0.657 $\pm$ 0.007 & 0.651 $\pm$ 0.008 & \textbf{0.353} $\pm$ 0.012 & 0.317 $\pm$ 0.007 & 0.287 $\pm$ 0.004 & 0.257 $\pm$ 0.004\\
%TMNtitle & w2v-\textsc{dmm} & 0.583 $\pm$ 0.020 & 0.665 $\pm$ 0.012 & 0.674  $\pm$  0.006 &  \textbf{0.681} $\pm$ 0.003 & 0.324 $\pm$ 0.013 & 0.329 $\pm$ 0.007 & 0.300  $\pm$  0.003 & 0.277 $\pm$ 0.003\\
%& glove-\textsc{dmm} &  \textbf{0.601} $\pm$ 0.021 & \textbf{0.670} $\pm$ 0.016  & \textbf{0.680} $\pm$ 0.005 & 0.679 $\pm$ 0.008 & \textbf{0.354} $\pm$ 0.013 & \textbf{0.333} $\pm$ 0.009 & \textbf{0.301} $\pm$ 0.003 & \textbf{0.278} $\pm$ 0.003\\
%\cline{2-10}
%& Improve. &  0.003 & 0.02 & 0.023 & 0.03 & 0.001 & 0.016 & 0.014 & 0.021\\

 & \textsc{dmm} & 0.558 $\pm$ 0.015 & 0.600 $\pm$ 0.010 & 0.634 $\pm$ 0.011 & 0.658 $\pm$ 0.006 & 0.338 $\pm$ 0.012 & 0.327 $\pm$ 0.006 & 0.304 $\pm$ 0.004 & 0.271 $\pm$ 0.002\\
TMNtitle & w2v-\textsc{dmm} & 0.552  $\pm$  0.022 & 0.653  $\pm$  0.012 & 0.678  $\pm$  0.007 & 0.682  $\pm$  0.005 & 0.314  $\pm$ 0.016 & 0.325  $\pm$  0.006 & 0.305  $\pm$  0.004 & \textbf{0.282}  $\pm$  0.003\\
& glove-\textsc{dmm} & \textbf{0.586}  $\pm$  0.019 & \textbf{0.672}  $\pm$  0.013 & \textbf{0.679}  $\pm$  0.009 & \textbf{0.683}  $\pm$  0.004 & \textbf{0.343}  $\pm$  0.015 & \textbf{0.339} $\pm$  0.007 & \textbf{0.307}  $\pm$  0.004 & \textbf{0.282}  $\pm$  0.002\\
\cline{2-10}
& Improve. & 0.028 & 0.072 & 0.045 & 0.025 & 0.005 & 0.012 & 0.003 & 0.011 \\

\hline 
\end{tabular}
%}
}
\caption{Purity and \textsc{nmi} results on the TMN and TMNtitle datasets with the mixture weight $\lambda = 0.6$.}
%\vspace{-10pt}
\label{tab:puritynmi1}
\end{table*}

\begin{table*}[!ht]
\centering
%{\scriptsize
\resizebox{16.5cm}{!}{
\def\arraystretch{0.975}
\setlength{\tabcolsep}{0.2em}
\begin{tabular}{l|l|l|l|l|l|l|l|l|l}
\hline
\multirow{2}{*}{{ Data}} & \multirow{2}{*}{Method}& \multicolumn{4}{|c}{Purity} & \multicolumn{4}{|c}{NMI}\\
\cline{3-10}
 & & T=4 & T=20 &  T=40 & T=80 & T=4 & T=20 &  T=40 & T=80 \\
\hline
\hline
 & \textsc{lda} & 0.559 $\pm$ 0.020 & 0.614 $\pm$ 0.016 & 0.626 $\pm$ 0.011 & 0.631 $\pm$ 0.008 & 0.196 $\pm$ 0.018 & 0.174 $\pm$ 0.008 & 0.170 $\pm$ 0.007 & 0.160 $\pm$ 0.004\\
 
Twitter & w2v-\textsc{lda} & \textbf{0.598} $\pm$ 0.023 & \textbf{0.635} $\pm$ 0.016 & \textbf{0.638} $\pm$ 0.009 & \textbf{0.637} $\pm$ 0.012 & \textbf{0.249} $\pm$ 0.021 & \textbf{0.191} $\pm$ 0.011 & 0.176 $\pm$ 0.003 & \textbf{0.167} $\pm$ 0.006\\

 & glove-\textsc{lda} & 0.597 $\pm$ 0.016 & \textbf{0.635} $\pm$ 0.014 & 0.637 $\pm$ 0.010 & \textbf{0.637} $\pm$ 0.007 & 0.242 $\pm$ 0.013 & \textbf{0.191} $\pm$ 0.007 & \textbf{0.177} $\pm$ 0.007 & 0.165 $\pm$ 0.005\\
\cline{2-10}
& Improve. & 0.039 & 0.021 & 0.012 & 0.006 & 0.053 & 0.017  & 0.007 & 0.007\\
\hline
%  & \textsc{dmm} & 0.552 $\pm$ 0.020 & 0.624 $\pm$ 0.010 & 0.647 $\pm$ 0.009 & 0.675 $\pm$ 0.009 & 0.194 $\pm$ 0.017 & 0.186 $\pm$ 0.006 & 0.184 $\pm$ 0.005 & 0.190 $\pm$ 0.003\\   
%Twitter & w2v-\textsc{dmm} & \textbf{0.581} $\pm$ 0.019 & 0.641 $\pm$ 0.013 & \textbf{0.660} $\pm$ 0.010 & \textbf{0.687} $\pm$ 0.007 & 0.230 $\pm$ 0.015 & 0.195 $\pm$ 0.007 & \textbf{0.193} $\pm$ 0.004 & \textbf{0.199} $\pm$ 0.005 \\
%    & glove-\textsc{dmm} & 0.580 $\pm$ 0.013 & \textbf{0.644} $\pm$ 0.016 & 0.657 $\pm$ 0.008 & 0.684 $\pm$ 0.006 &  \textbf{0.232} $\pm$ 0.010 & \textbf{0.201} $\pm$ 0.010 & 0.191 $\pm$ 0.006 & 0.195 $\pm$ 0.005 \\
%\cline{2-10}
%& Improve. &  0.029 & 0.02 & 0.013 & 0.012 & 0.038 & 0.015 & 0.009 & 0.009 \\

  & \textsc{dmm} & 0.523 $\pm$ 0.011 & 0.619 $\pm$ 0.015 & 0.660 $\pm$ 0.008 & 0.684 $\pm$ 0.010 & 0.222 $\pm$ 0.013 & 0.213 $\pm$ 0.011 & 0.198 $\pm$ 0.008 & 0.196 $\pm$ 0.004\\   
Twitter & w2v-\textsc{dmm} & \textbf{0.589}  $\pm$ 0.017 & 0.655  $\pm$  0.015 & \textbf{0.668}  $\pm$  0.008 & 0.694  $\pm$  0.009 & 0.243  $\pm$  0.014 & 0.215  $\pm$  0.009 & \textbf{0.203}  $\pm$  0.005 & 0.204  $\pm$  0.006\\
    & glove-\textsc{dmm} &  0.583  $\pm$  0.023 & \textbf{0.661}  $\pm$  0.019 & 0.667  $\pm$  0.009 & \textbf{0.697}  $\pm$  0.009 & \textbf{0.250}  $\pm$  0.020 & \textbf{0.223}  $\pm$  0.014 & 0.201  $\pm$  0.006 & \textbf{0.206}  $\pm$  0.005\\
\cline{2-10}
& Improve. & 0.066 & 0.042 & 0.008 & 0.013 & 0.028 & 0.01 & 0.005 & 0.01   \\

\hline
\end{tabular}
%}
}
\caption{Purity and \textsc{nmi} results on the Twitter dataset with the mixture weight $\lambda = 0.6$.}
\label{tab:puritynmi2}
\end{table*}

Figures \ref{fig:N20short20T} and \ref{fig:N20short40T} present Purity and \textsc{nmi} results obtained by the \textsc{lda}, w2v-\textsc{lda} and glove-\textsc{lda} models on the N20short dataset with the numbers of topics $T$ set to either 20 or 40,  and the value of the mixture weight $\lambda$  varied from 0.0 to 1.0. 

We found that setting $\lambda$ to 1.0 (i.e.  using only the latent features to model words), the glove-\textsc{lda} produced  1\%+ higher scores on both Purity and \textsc{nmi} results than the w2v-\textsc{lda} when using $20$ topics. However, the two models glove-\textsc{lda} and w2v-\textsc{lda} returned equivalent results with $40$ topics where they gain 2\%+ absolute improvement\footnote{Using the Student's t-Test, the improvement is significant ($p < 0.01$).} on the two Purity and \textsc{nmi} against the baseline \textsc{lda} model. 

By varying  $\lambda$, as shown in Figures \ref{fig:N20short20T} and \ref{fig:N20short40T}, the w2v-\textsc{lda} and glove-\textsc{lda} models obtain their best results at $\lambda=0.6$  where the w2v-\textsc{lda} model does slightly better than the glove-\textsc{lda}. Both models significantly outperform their baseline \textsc{lda} models; for example with $40$ topics, the w2v-\textsc{lda} model attains 4.4\% and 4.3\% over the \textsc{lda} model on  Purity and \textsc{nmi} metrics, respectively. 

We fix the mixture weight $\lambda$  at 0.6, and report experimental results based on this value \CHANGEA{for} the rest of this section. 
 Tables \ref{tab:puritynmi}, \ref{tab:puritynmi1} and \ref{tab:puritynmi2} show clustering results produced by our  models and the baseline models on the remaining datasets with different numbers of topics.  As expected, the \textsc{dmm} model is better than the \textsc{lda} model  on the short datasets of TMN, TMNtitle and Twitter. \CHANGEC{ For example with 80 topics on the TMNtitle dataset, the \textsc{dmm}  achieves about 7+\% higher Purity and \textsc{nmi} scores than  \textsc{lda}}. %In the other short datasets TMN and Twitter, the \textsc{dmm} model produces results competitive with the \textsc{lda} model at a small value of $T$, for example $T \leq 7$, while it obtains better results than the \textsc{lda} model with larger values $T$.

%We compare the highest results gained by our  models with the baseline results in section \ref{ssub:mixvsbl} and examine  the difference between using Google Word2Vec  and Stanford GloVe word vectors in section \ref{ssub:glvsw2v}.

%\subsubsection{Our models vs. baseline models}\label{ssub:mixvsbl}

\textbf{New models vs. baseline models:}  
On most tests, our  models score higher than the baseline models, particularly on the small N20small dataset  where we get 6.0\% improvement on \textsc{nmi} at $T=6$, and on the short text  TMN and TMNtitle datasets we  obtain \CHANGEC{ 6.1\% and 2.5\%} higher Purity at $T=80$. In addition,  on the short and small Twitter dataset with $T=4$,  we achieve 3.9\% and 5.3\% improvements in  Purity and \textsc{nmi} scores, respectively. Those results show that an improved model of topic-word mappings also improves \CHANGEA{the} document-topic assignments.

%On most tests, our combined models surpass the baseline models, and particularly achieve significantly  higher results against the baseline on the small dataset N20small where we get 7.1\% improvement on \textsc{nmi}, and on short text dataset TMN where we obtain 4.8\% higher result for Purity. 

For  the small value of $T \leq 7$, on the large datasets of N20, TMN and TMNtitle, our  models and baseline models obtain similar  clustering results. %\footnote{The student's t-Test shows the difference between the best result obtained by our  model and the baseline model is not significant.}. 
 However, with higher values of $T$, our models perform  better than the baselines on the short TMN and TMNtitle datasets, while on the N20 dataset, the baseline \textsc{lda} model attains a slightly higher clustering results than ours. In contrast, on the short and small Twitter dataset, our models obtain considerably better clustering results than the baseline models with a small value of $T$.%  (the results for our models and the baseline models are similar for larger values of $T$).

\textbf{Google word2vec vs. Stanford glove word vectors:}  \CHANGEA{On the small N20short and N20small datasets}, using the Google pre-trained word vectors produces higher clustering \CHANGEA{scores} than using Stanford pre-trained word vectors. However, on \CHANGEA{the} large datasets N20, TMN and TMNtitle, using Stanford word vectors produces  higher scores than using Google word vectors when using a smaller number of topics, for example $T \leq 20$. With more topics, for instance $T=80$, the pre-trained Google and Stanford word vectors produce similar clustering results. In addition, on the Twitter dataset, both sets of pre-trained word vectors produce similar results.% \textbf{REASON why???}

\subsection{Document classification evaluation}

%This section compares our models with the baseline models in a document classification task. 
Unlike the document clustering task, \CHANGEA{the} document classification task evaluates the
distribution over topics for each document. Following \newcite{NIPS2008_3599}, \newcite{LuMZ2011}, \newcite{Huh2012} and
\newcite{ICML2013_zhai13}, we used
Support Vector Machines (SVM)
  % algorithm  on the topic-proportion vectors of documents
to predict
the ground truth labels from the topic-proportion vector of each document.  We \CHANGEA{used} the WEKA's  implementation \cite{Hall:2009} of
  % SVM's fast training
the fast Sequential Minimal Optimization algorithm \cite{Platt:1999} for \CHANGEA{learning a classifier
with ten-fold cross-validation}
%\footnote{We performed a 10-fold cross-validation evaluation scheme for each of ten repetitions of every experiment.} 
 and WEKA's default parameters. We present the macro-averaged  $F_1$ score \cite[Section 13.6]{ManningRS2008} as the evaluation metric for this task.
 % Higher F-measure scores reflect better classification performance.

%\begin{figure}[!ht]
%\centering
%\includegraphics[width=8cm,height=4.75cm]{figures/ClassificationN20short20T.png}
%\caption{ $F_1$ scores (mean and standard deviation)   on the N20short dataset with number of topics $T$ = 20, varying $\lambda$ from 0.0 to 1.0.}
%\label{fig:ClassificationN20short20T}
%\end{figure}
%%\vspace{-10pt}
%
% \begin{figure}[!ht]
%\centering
%\includegraphics[width=8cm,height=4.75cm]{figures/ClassificationN20short40T.png}
%\caption{$F_1$ scores  on the N20short dataset with number of topics $T=40$, varying  $\lambda$ from 0.0 to 1.0.}
%\label{fig:ClassificationN20short40T}
%\end{figure}

%Figures \ref{fig:ClassificationN20short20T} and \ref{fig:ClassificationN20short40T} present $F_1$ scores achieved by the \textsc{lda}, w2v-\textsc{lda} and glove-\textsc{lda} models on the N20short dataset. 

\CHANGEA{Just as in} the document clustering task, the mixture weight $\lambda = 0.6$ obtains the highest classification performances on the N20short dataset. For example with $T=40$, \CHANGEA{our w2v-\textsc{lda} and glove-\textsc{lda}  obtain $F_1$ scores at 40.0\% and 38.9\% which are 4.5\% and 3.4\%  higher than   $F_1$ score at 35.5\% obtained by the  \textsc{lda} model, respectively}.

We report classification results on the remaining experimental datasets with mixture weight $\lambda = 0.6$ in tables \ref{tab:f1score}, \ref{tab:f1score2} and \ref{tab:f1score3}. \CHANGEA{Unlike the} clustering results, the \textsc{lda} model does better than the \textsc{dmm} model for classification on the TMN dataset. %Additionally, the \textsc{lda} model also returns better classification scores than the \textsc{dmm} model on the Twitter dataset with small values of $T$.

\begin{table}[!ht]
\centering
\setlength{\tabcolsep}{0.2em}
\resizebox{8.0cm}{!}{
\begin{tabular}{l|l|l|l|l|l}
\hline
\multirow{2}{*}{{ Data}} & \multirow{2}{*}{Method}&  \multicolumn{3}{|c}{$\lambda = 0.6$}\\
\cline{3-6}
&  & T=6 & T=20 &  T=40 & T=80 \\
\hline
\hline
 & \textsc{lda} &  0.312 $\pm$ 0.013 & 0.635  $\pm$ 0.016 & \textbf{0.742} $\pm$ 0.014 & 0.763 $\pm$ 0.005 \\
 
N20 & w2v-\textsc{lda} & \textbf{0.316} $\pm$ 0.013 & 0.641 $\pm$ 0.019 & 0.730 $\pm$ 0.017 & \textbf{0.768} $\pm$ 0.004 \\

 & glove-\textsc{lda} & 0.288 $\pm$ 0.013 & \textbf{0.650} $\pm$ 0.024 & 0.733 $\pm$ 0.011 & 0.762 $\pm$ 0.006\\
\cline{2-6}
& Improve.  & 0.004 & 0.015 & -0.009 & 0.005 \\
\hline
\hline
 & \textsc{lda} & 0.204 $\pm$ 0.020 & 0.392 $\pm$ 0.029 & 0.459 $\pm$ 0.030 & 0.477 $\pm$ 0.025 \\
N20small & w2v-\textsc{lda} &  \textbf{0.213} $\pm$ 0.018 & \textbf{0.442} $\pm$ 0.025 & \textbf{0.502} $\pm$ 0.031 & \textbf{0.509} $\pm$ 0.022\\
 & glove-\textsc{lda} &  0.181 $\pm$ 0.011 & 0.420 $\pm$ 0.025 & 0.474 $\pm$ 0.029 & 0.498 $\pm$ 0.012\\
\cline{2-6}
& Improve. &  0.009 & 0.05 & 0.043 & 0.032 \\
\hline
\end{tabular}
}
\caption{$F_1$ scores (mean and standard deviation) for N20 and N20small datasets.}% \textit{Improve.} row denotes is the difference between the best result obtained by our model and the baseline model.}
\label{tab:f1score}
\end{table}

%%\vspace{-10pt}

\begin{table}[!ht]
\centering
\setlength{\tabcolsep}{0.2em}
\resizebox{8.0cm}{!}{
\begin{tabular}{l|l|l|l|l|l}
\hline
\multirow{2}{*}{{ Data}} & \multirow{2}{*}{Method}&  \multicolumn{4}{|c}{$\lambda = 0.6$}\\
\cline{3-6}
&  & T=7 & T=20 &  T=40 & T=80 \\
\hline
\hline
 & \textsc{lda}& 0.658 $\pm$ 0.026 & 0.754 $\pm$ 0.009 & 0.768 $\pm$ 0.004 & 0.778 $\pm$ 0.004\\
TMN & w2v-\textsc{lda} & 0.663 $\pm$ 0.021 & 0.758 $\pm$ 0.009 & \textbf{0.769} $\pm$ 0.005 & \textbf{0.780} $\pm$ 0.004 \\
& glove-\textsc{lda} & \textbf{0.664} $\pm$ 0.025 & \textbf{0.760} $\pm$ 0.006 & 0.767 $\pm$ 0.003 & 0.779 $\pm$ 0.004 \\
\cline{2-6}
& Improve. & 0.006 & 0.006 & 0.001 & 0.002 \\
\hline
% & \textsc{dmm} &  0.605 $\pm$ 0.023 & 0.724 $\pm$ 0.016 & 0.738 $\pm$ 0.008 & 0.741 $\pm$ 0.005  \\
%TMN & w2v-\textsc{dmm} & 0.619 $\pm$ 0.033 & 0.744 $\pm$ 0.009 & 0.759 $\pm$ 0.005 & \textbf{0.777} $\pm$ 0.005\\
% & glove-\textsc{dmm} & \textbf{0.624} $\pm$ 0.025 & \textbf{0.757} $\pm$ 0.009 & \textbf{0.761} $\pm$  0.005 & 0.774 $\pm$ 0.010\\
%\cline{2-6}
%& Improve. &  0.019 & 0.033 & 0.023 & 0.036\\

 & \textsc{dmm} &  0.607 $\pm$ 0.040 & 0.694  $\pm$  0.026 & 0.712  $\pm$  0.014 & 0.721  $\pm$  0.008  \\
TMN & w2v-\textsc{dmm} & 0.607  $\pm$  0.019 & 0.736  $\pm$  0.025 & \textbf{0.760}  $\pm$  0.011 & 0.771 $\pm$  0.005 \\
 & glove-\textsc{dmm} &  \textbf{0.621}  $\pm$  0.042 & \textbf{0.750}  $\pm$  0.011 & 0.759  $\pm$  0.006 & \textbf{0.775}  $\pm$  0.006\\
\cline{2-6}
& Improve. & 0.014 & 0.056 & 0.048 & 0.054 \\

\hline
\hline
 & \textsc{lda} &  0.564 $\pm$ 0.015 & 0.625 $\pm$ 0.011 & 0.626 $\pm$ 0.010 & 0.624 $\pm$ 0.006 \\
TMNtitle & w2v-\textsc{lda} & 0.563 $\pm$ 0.029 & \textbf{0.644} $\pm$ 0.010 & \textbf{0.643} $\pm$ 0.007 & 0.640 $\pm$ 0.004 \\
& glove-\textsc{lda} &  \textbf{0.568} $\pm$ 0.028 & \textbf{0.644} $\pm$ 0.010 & 0.632 $\pm$ 0.008 & \textbf{0.642} $\pm$ 0.005 \\
\cline{2-6}
& Improve. & 0.004 & 0.019 & 0.017 & 0.018  \\
\hline
% & \textsc{dmm} &  0.570 $\pm$ 0.022 & 0.650 $\pm$ 0.011 & 0.654 $\pm$ 0.008 & 0.646 $\pm$ 0.008 \\
%TMNtitle & w2v-\textsc{dmm} &  0.562 $\pm$ 0.022 & 0.670 $\pm$ 0.012 & 0.677 $\pm$ 0.006 & \textbf{0.680} $\pm$ 0.003\\
%& glove-\textsc{dmm}  &  \textbf{0.592} $\pm$ 0.017 & \textbf{0.674} $\pm$ 0.016 & \textbf{0.683} $\pm$ 0.006 & 0.679 $\pm$ 0.009\\
%\cline{2-6}
%& Improve. & 0.022 & 0.024 & 0.029 & 0.034 \\

 & \textsc{dmm} & 0.500 $\pm$ 0.021 & 0.600 $\pm$ 0.015 & 0.630 $\pm$ 0.016 & 0.652 $\pm$ 0.005   \\
TMNtitle & w2v-\textsc{dmm} & 0.528 $\pm$ 0.028 & 0.663 $\pm$ 0.008 & 0.682 $\pm$ 0.008 & \textbf{0.681} $\pm$ 0.006  \\
& glove-\textsc{dmm}  &   \textbf{0.565} $\pm$ 0.022 & \textbf{0.680} $\pm$ 0.011 & \textbf{0.684} $\pm$ 0.009 & \textbf{0.681} $\pm$ 0.004\\
\cline{2-6}
& Improve. &  0.065 & 0.08 & 0.054 & 0.029  \\
\hline 
\end{tabular}
}
\caption{$F_1$ scores for TMN  and TMNtitle datasets.}% \textit{Improve.} denotes the absolute  improvement accounted for the best result (in bold) produced by our latent feature model over the baseline models.}
\label{tab:f1score2}
\end{table}

%%\vspace{-10pt}

\begin{table}[!ht]
\centering
\setlength{\tabcolsep}{0.2em}
\resizebox{8.0cm}{!}{
\begin{tabular}{l|l|l|l|l|l}
\hline
\multirow{2}{*}{{ Data}} & \multirow{2}{*}{Method}&  \multicolumn{4}{|c}{$\lambda = 0.6$}\\
\cline{3-6}
&  & T=4 & T=20 &  T=40 & T=80 \\
\hline
\hline
 & \textsc{lda} & 0.526 $\pm$ 0.021 & 0.636 $\pm$ 0.011 & 0.650 $\pm$ 0.014 & 0.653 $\pm$ 0.008  \\
Twitter & w2v-\textsc{lda} & \textbf{0.578}  $\pm$ 0.047 & 0.651 $\pm$ 0.015 & 0.661 $\pm$ 0.011 & \textbf{0.664} $\pm$ 0.010  \\
& glove-\textsc{lda} & 0.569 $\pm$ 0.037 & \textbf{0.656} $\pm$ 0.011 & \textbf{0.662} $\pm$ 0.008 & 0.662 $\pm$ 0.006\\
\cline{2-6}
& Improve.  & 0.052 & 0.02 & 0.012 & 0.011 \\
\hline
% & \textsc{dmm} & 0.505 $\pm$ 0.023 & 0.614 $\pm$ 0.012 & 0.634 $\pm$ 0.013 & 0.656 $\pm$ 0.011 \\
%Twitter & w2v-\textsc{dmm} & \textbf{0.541} $\pm$ 0.035 & 0.636 $\pm$ 0.015 & \textbf{0.648} $\pm$ 0.011 & \textbf{0.670} $\pm$ 0.010  \\
% & glove-\textsc{dmm} &  0.539 $\pm$  0.024  & \textbf{0.638} $\pm$ 0.017 & 0.645 $\pm$ 0.012 & 0.666 $\pm$ 0.009\\
%\cline{2-6}
%& Improve.  &  0.036 & 0.024 & 0.014 & 0.014  \\

 & \textsc{dmm} &  0.469 $\pm$ 0.014 & 0.600 $\pm$ 0.021 & 0.645 $\pm$ 0.009 & 0.665 $\pm$ 0.014 \\
Twitter & w2v-\textsc{dmm} & \textbf{0.539} $\pm$ 0.016 & 0.649 $\pm$ 0.016 & 0.656 $\pm$ 0.007 & 0.676 $\pm$ 0.012 \\
 & glove-\textsc{dmm} & 0.536 $\pm$ 0.027 & \textbf{0.654} $\pm$ 0.019 & \textbf{0.657} $\pm$ 0.008 & \textbf{0.680} $\pm$ 0.009 \\
\cline{2-6}
& Improve.  & 0.07 & 0.054 & 0.012 & 0.015    \\
\hline
\end{tabular}
}
\caption{$F_1$ scores for Twitter dataset.}% \textit{Improve.} denotes the absolute  improvement accounted for the best result (in bold) produced by our latent feature model over the baseline models.}
\label{tab:f1score3}
\end{table}

\textbf{New models vs. baseline models:} On most evaluations, our models perform better than the baseline
models. In particular, on the small N20small and Twitter datasets, when the number of topics $T$ is
equal to number of ground truth labels (i.e. 20 and 4 correspondingly), our w2v-\textsc{lda} obtains $5^+$\%  higher $F_1$ score than the  \textsc{lda} model. In addition,  our w2v-\textsc{dmm} model  achieves \CHANGEC{ 5.4\% and 2.9\% } higher $F_1$ score than the  \textsc{dmm} model on short  TMN and TMNtitle datasets with $T=80$, respectively. 

% On the TMNtitle dataset with a small value  $T=7$, unlike in the clustering task where the glove-\textsc{dmm} and \textsc{dmm} obtain  similar clustering results, the glove-\textsc{dmm} model scores higher than the \textsc{dmm} model with 2.2\%  absolute classification improvement\footnote{Using the student's t-Test, the improvement is significant with a p-value less than 0.05.}.

\textbf{Google word2vec vs. Stanford glove word vectors:} The comparison \CHANGEA{of the} Google and Stanford pre-trained word vectors for  classification  is similar to the one for  clustering. %For example, using Google   vectors produces better classification performances than using  Stanford  vectors on the small datasets N20short, N20small and Twitter.

\subsection{Discussion}

\CHANGEA{We found that the topic coherence evaluation produced the best results with a mixture weight $\lambda=1$, which corresponds to using topic-word distributions defined in terms of the latent-feature word vectors.  This is not surprising, since the topic coherence evaluation we used \cite{Lau2014} is based on word co-occurrences in an external corpus (here, Wikipedia), and it is reasonable that the billion-word corpora used to train the latent feature word vectors are more useful for this task than the much smaller topic-modeling corpora, from which \CHANGEB{the topic-word multinomial distributions} are trained}.

\CHANGEA{On the other hand, the document clustering and  document classification tasks depend more strongly on possibly idiosyncratic properties of the smaller topic-modeling corpora, since these evaluations reflect how well the document-topic assignments can group or distinguish documents within the topic-modeling corpus.  Smaller values of $\lambda$ enable the models to learn topic-word distributions that include an arbitrary \CHANGEB{multinomial} topic-word distribution, enabling the models to capture idiosyncratic properties of the topic-modeling corpus.  Even in these evaluations we found that an intermediate value of
$\lambda=0.6$ produced the best results, indicating that better word-topic distributions were produced when information from the large external corpus is combined with corpus-specific topic-word multinomials.  We found that using the latent feature word vectors produced significant performance improvements even when the domain of the topic-modeling corpus was quite different to that of the external corpus from which the word vectors were derived, as was the case in our experiments on Twitter data}.

\CHANGEA{We found that using either the Google or the Stanford latent feature word vectors produced very similar results.  As far as we could tell, there is no reason to prefer either one of these in our topic modeling applications}.

\section{Conclusion and future work}

In this paper, we have shown that latent feature representations can be used to improve topic models. We proposed two novel latent feature topic models, namely \textsc{lf-lda} and \textsc{lf-dmm}, that integrate a latent feature model within two topic models \textsc{lda} and \textsc{dmm}. 
We compared the performance of our models  \textsc{lf-lda} and \textsc{lf-dmm} to the baseline  \textsc{lda} and \textsc{dmm} models on  topic coherence, document clustering and document classification evaluations. In the topic coherence evaluation, our model outperformed the baseline models on all 6 experimental datasets, showing that our method for exploiting  external information from very large corpora helps improve the topic-to-word mapping. 
Meanwhile, document clustering and document classification results show that our models  \CHANGEA{improve the} document-topic assignments compared to the baseline models, especially on datasets with few  or short documents.

\CHANGEA{As an anonymous reviewer suggested, it would be interesting to identify exactly how the latent feature word vectors improve topic modeling performance.  We believe that they provide useful information about word meaning extracted from the large corpora that they are trained on, but as the reviewer suggested, it is possible that the performance improvements arise because the word vectors are trained on context windows of size 5 or 10, while the \textsc{lda} and \textsc{dmm} models view documents as bags of words, and effectively use a context window that encompasses the entire document.  In preliminary experiments where we train latent feature word vectors from the topic-modeling corpus alone using context windows of size 10 we found that performance was degraded relative to the results presented here, suggesting that the use of a context window alone is not responsible for the performance improvements we reported here.  Clearly it would be valuable to investigate this further}.

\CHANGEA{In order to use a Gibbs sampler in section \ref{gibbslfdmm}}, the conditional distributions needed to
be distributions we can sample from cheaply, which is not the case for the ratios of Gamma
functions. While we used a \CHANGEA{simple} approximation, it is worth exploring other sampling techniques that can
avoid approximations, such as Metropolis-Hastings sampling \cite[Section 11.2.2]{Bishop2006}.

In order to compare the pre-trained Google and Stanford word vectors, we excluded words that did not
appear in both sets of vectors. As suggested by anonymous reviewers, \CHANGEA{it would be interesting to learn  vectors for these unseen words. In addition, it is worth fine-tuning the seen-word vectors on the dataset of interest.}

Although we have not evaluated our approach on  very large corpora, the corpora we have evaluated
on do vary in size, and we \CHANGEA{showed} that the gains from our approach are greatest when the corpora are
small.  \CHANGEA{A drawback of our approach} is that it is slow on very large corpora. Variational Bayesian
inference may provide an efficient solution to this problem \cite{Jordan1999,Blei2003}.

\section*{Acknowledgments}  

This research was supported by a Google award through the Natural 
Language Understanding Focused Program, and under the Australian 
Research Council's {\em Discovery Projects} funding scheme (project 
numbers DP110102506 and DP110102593). The authors would like to thank the three anonymous reviewers, the action editor and  Dr. John Pate at the Macquarie University, Australia for helpful comments and suggestions.

\bibliographystyle{acl2012}
%\balance
\bibliography{References}

\begin{thebibliography}{}

\bibitem[\protect\citename{Bishop}2006]{Bishop2006}
Christopher~M. Bishop.
\newblock 2006.
\newblock {\em {Pattern Recognition and Machine Learning (Information Science
  and Statistics)}}.
\newblock Springer-Verlag New York, Inc.

\bibitem[\protect\citename{Blei \bgroup et al.\egroup }2003]{Blei2003}
David~M. Blei, Andrew~Y. Ng, and Michael~I. Jordan.
\newblock 2003.
\newblock {Latent Dirichlet Allocation}.
\newblock {\em Journal of Machine Learning Research}, 3:993--1022.

\bibitem[\protect\citename{Blei}2012]{Blei2012}
David~M. Blei.
\newblock 2012.
\newblock {Probabilistic Topic Models}.
\newblock {\em Communications of the ACM}, 55(4):77--84.

\bibitem[\protect\citename{Bullinaria and Levy}2007]{Bullinaria07}
John~A. Bullinaria and Joseph~P. Levy.
\newblock 2007.
\newblock {Extracting semantic representations from word co-occurrence
  statistics: A computational study}.
\newblock {\em Behavior Research Methods}, 39(3):510--526.

\bibitem[\protect\citename{Cai \bgroup et al.\egroup }2008]{Cai2008}
Deng Cai, Qiaozhu Mei, Jiawei Han, and Chengxiang Zhai.
\newblock 2008.
\newblock {Modeling Hidden Topics on Document Manifold}.
\newblock In {\em Proceedings of the 17th ACM Conference on Information and
  Knowledge Management}, pages 911--920.

\bibitem[\protect\citename{Cao \bgroup et al.\egroup }2015]{AAAI159303}
Ziqiang Cao, Sujian Li, Yang Liu, Wenjie Li, and Heng Ji.
\newblock 2015.
\newblock {A Novel Neural Topic Model and Its Supervised Extension}.
\newblock In {\em Proceedings of the Twenty-Ninth AAAI Conference on Artificial
  Intelligence}, pages 2210--2216.

\bibitem[\protect\citename{Cardoso-Cachopo}2007]{Ana-Cardoso-Cachopo}
Ana Cardoso-Cachopo.
\newblock 2007.
\newblock {Improving Methods for Single-label Text Categorization}.
\newblock PhD Thesis, Instituto Superior Tecnico, Universidade Tecnica de
  Lisboa.

\bibitem[\protect\citename{Chang \bgroup et al.\egroup }2009]{NIPS2009_3700}
Jonathan Chang, Sean Gerrish, Chong Wang, Jordan~L. Boyd-graber, and David~M.
  Blei.
\newblock 2009.
\newblock {Reading Tea Leaves: How Humans Interpret Topic Models}.
\newblock In {\em Advances in Neural Information Processing Systems 22}, pages
  288--296.

\bibitem[\protect\citename{Collobert and Weston}2008]{Collobert2008}
Ronan Collobert and Jason Weston.
\newblock 2008.
\newblock {A Unified Architecture for Natural Language Processing: Deep Neural
  Networks with Multitask Learning}.
\newblock In {\em Proceedings of the 25th International Conference on Machine
  Learning}, pages 160--167.

\bibitem[\protect\citename{Deerwester \bgroup et al.\egroup }1990]{deer1990lsa}
Scott Deerwester, Susan~T. Dumais, George~W. Furnas, Thomas~K. Landauer, and
  Richard Harshman.
\newblock 1990.
\newblock {Indexing by Latent Semantic Analysis}.
\newblock {\em Journal of the American Society for Information Science},
  41(6):391--407.

\bibitem[\protect\citename{Eisenstein \bgroup et al.\egroup
  }2011]{ICML2011Eisenstein_534}
Jacob Eisenstein, Amr Ahmed, and Eric Xing.
\newblock 2011.
\newblock {Sparse Additive Generative Models of Text}.
\newblock In {\em Proceedings of the 28th International Conference on Machine
  Learning}, pages 1041--1048.

\bibitem[\protect\citename{Glorot \bgroup et al.\egroup
  }2011]{ICML2011Glorot_342}
Xavier Glorot, Antoine Bordes, and Yoshua Bengio.
\newblock 2011.
\newblock {Domain Adaptation for Large-Scale Sentiment Classification: A Deep
  Learning Approach}.
\newblock In {\em Proceedings of the 28th International Conference on Machine
  Learning}, pages 513--520.

\bibitem[\protect\citename{Griffiths and Steyvers}2004]{GriffithsS2004}
Thomas~L. Griffiths and Mark Steyvers.
\newblock 2004.
\newblock {Finding scientific topics}.
\newblock {\em Proceedings of the National Academy of Sciences of the United
  States of America}, 101(Suppl 1):5228--5235.

\bibitem[\protect\citename{Hall \bgroup et al.\egroup }2009]{Hall:2009}
Mark Hall, Eibe Frank, Geoffrey Holmes, Bernhard Pfahringer, Peter Reutemann,
  and Ian~H. Witten.
\newblock 2009.
\newblock {The WEKA Data Mining Software: An Update}.
\newblock {\em ACM SIGKDD Explorations Newsletter}, 11(1):10--18.

\bibitem[\protect\citename{Han \bgroup et al.\egroup }2012]{han2012}
Bo~Han, Paul Cook, and Timothy Baldwin.
\newblock 2012.
\newblock {Automatically Constructing a Normalisation Dictionary for
  Microblogs}.
\newblock In {\em Proceedings of the 2012 Joint Conference on Empirical Methods
  in Natural Language Processing and Computational Natural Language Learning},
  pages 421--432.

\bibitem[\protect\citename{Hingmire \bgroup et al.\egroup
  }2013]{Hingmire:2013:DCT:2484028.2484140}
Swapnil Hingmire, Sandeep Chougule, Girish~K. Palshikar, and Sutanu
  Chakraborti.
\newblock 2013.
\newblock {Document Classification by Topic Labeling}.
\newblock In {\em Proceedings of the 36th international ACM SIGIR conference on
  Research and development in information retrieval}, pages 877--880.

\bibitem[\protect\citename{Hong and Davison}2010]{Hong2010D}
Liangjie Hong and Brian~D. Davison.
\newblock 2010.
\newblock {Empirical Study of Topic Modeling in Twitter}.
\newblock In {\em Proceedings of the First Workshop on Social Media Analytics},
  pages 80--88.

\bibitem[\protect\citename{Huh and Fienberg}2012]{Huh2012}
Seungil Huh and Stephen~E. Fienberg.
\newblock 2012.
\newblock {Discriminative Topic Modeling Based on Manifold Learning}.
\newblock {\em ACM Transactions on Knowledge Discovery from Data},
  5(4):20:1--20:25.

\bibitem[\protect\citename{Johnson}2010]{johnson:2010:ACL}
Mark Johnson.
\newblock 2010.
\newblock {PCFGs, Topic Models, Adaptor Grammars and Learning Topical
  Collocations and the Structure of Proper Names}.
\newblock In {\em Proceedings of the 48th Annual Meeting of the Association for
  Computational Linguistics}, pages 1148--1157.

\bibitem[\protect\citename{Jordan \bgroup et al.\egroup }1999]{Jordan1999}
Michael~I. Jordan, Zoubin Ghahramani, Tommi~S. Jaakkola, and Lawrence~K. Saul.
\newblock 1999.
\newblock {An Introduction to Variational Methods for Graphical Models}.
\newblock {\em Machine Learning}, 37(2):183--233.

\bibitem[\protect\citename{Lacoste-Julien \bgroup et al.\egroup
  }2009]{NIPS2008_3599}
Simon Lacoste-Julien, Fei Sha, and Michael~I. Jordan.
\newblock 2009.
\newblock {DiscLDA: Discriminative Learning for Dimensionality Reduction and
  Classification}.
\newblock In {\em Advances in Neural Information Processing Systems 21}, pages
  897--904.

\bibitem[\protect\citename{Lau \bgroup et al.\egroup }2014]{Lau2014}
Han~Jey Lau, David Newman, and Timothy Baldwin.
\newblock 2014.
\newblock {Machine Reading Tea Leaves: Automatically Evaluating Topic Coherence
  and Topic Model Quality}.
\newblock In {\em Proceedings of the 14th Conference of the European Chapter of
  the Association for Computational Linguistics}, pages 530--539.

\bibitem[\protect\citename{Liu and Nocedal}1989]{Liu1989}
D.~C. Liu and J.~Nocedal.
\newblock 1989.
\newblock {On the Limited Memory BFGS Method for Large Scale Optimization}.
\newblock {\em Mathematical Programming}, 45(3):503--528.

\bibitem[\protect\citename{Liu \bgroup et al.\egroup }2015]{AAAI159314}
Yang Liu, Zhiyuan Liu, Tat-Seng Chua, and Maosong Sun.
\newblock 2015.
\newblock {Topical Word Embeddings}.
\newblock In {\em Proceedings of the Twenty-Ninth AAAI Conference on Artificial
  Intelligence}, pages 2418--2424.

\bibitem[\protect\citename{Lu \bgroup et al.\egroup }2011]{LuMZ2011}
Yue Lu, Qiaozhu Mei, and ChengXiang Zhai.
\newblock 2011.
\newblock {Investigating task performance of probabilistic topic models: an
  empirical study of PLSA and LDA}.
\newblock {\em Information Retrieval}, 14:178--203.

\bibitem[\protect\citename{Lund and Burgess}1996]{Lund96}
Kevin Lund and Curt Burgess.
\newblock 1996.
\newblock {Producing high-dimensional semantic spaces from lexical
  co-occurrence}.
\newblock {\em Behavior Research Methods, Instruments, \& Computers},
  28(2):203--208.

\bibitem[\protect\citename{Manning \bgroup et al.\egroup }2008]{ManningRS2008}
Christopher~D. Manning, Prabhakar Raghavan, and Hinrich Sch\"{u}tze.
\newblock 2008.
\newblock {\em {Introduction to Information Retrieval}}.
\newblock Cambridge University Press.

\bibitem[\protect\citename{McCallum}2002]{AndrewMcCallum2002}
Andrew McCallum.
\newblock 2002.
\newblock {MALLET: A Machine Learning for Language Toolkit}.

\bibitem[\protect\citename{Mehrotra \bgroup et al.\egroup }2013]{Mehrotra2013}
Rishabh Mehrotra, Scott Sanner, Wray Buntine, and Lexing Xie.
\newblock 2013.
\newblock {Improving LDA Topic Models for Microblogs via Tweet Pooling and
  Automatic Labeling}.
\newblock In {\em Proceedings of the 36th International ACM SIGIR Conference on
  Research and Development in Information Retrieval}, pages 889--892.

\bibitem[\protect\citename{Mikolov \bgroup et al.\egroup
  }2013]{MikolovNIPS2013}
Tomas Mikolov, Ilya Sutskever, Kai Chen, Greg~S Corrado, and Jeff Dean.
\newblock 2013.
\newblock {Distributed Representations of Words and Phrases and their
  Compositionality}.
\newblock In {\em Advances in Neural Information Processing Systems 26}, pages
  3111--3119.

\bibitem[\protect\citename{Mimno \bgroup et al.\egroup }2011]{Mimno2011}
David Mimno, Hanna~M. Wallach, Edmund Talley, Miriam Leenders, and Andrew
  McCallum.
\newblock 2011.
\newblock {Optimizing Semantic Coherence in Topic Models}.
\newblock In {\em Proceedings of the 2011 Conference on Empirical Methods in
  Natural Language Processing}, pages 262--272.

\bibitem[\protect\citename{Newman \bgroup et al.\egroup
  }2006]{Newman:2006:SEM:1150402.1150487}
David Newman, Chaitanya Chemudugunta, and Padhraic Smyth.
\newblock 2006.
\newblock {Statistical Entity-Topic Models}.
\newblock In {\em Proceedings of the 12th ACM SIGKDD international conference
  on Knowledge discovery and data mining}, pages 680--686.

\bibitem[\protect\citename{Newman \bgroup et al.\egroup }2010]{Newman2010}
David Newman, Jey~Han Lau, Karl Grieser, and Timothy Baldwin.
\newblock 2010.
\newblock {Automatic Evaluation of Topic Coherence}.
\newblock In {\em Human Language Technologies: The 2010 Annual Conference of
  the North American Chapter of the Association for Computational Linguistics},
  pages 100--108.

\bibitem[\protect\citename{Nigam \bgroup et al.\egroup }2000]{Nigam2000}
Kamal Nigam, AK~McCallum, S~Thrun, and T~Mitchell.
\newblock 2000.
\newblock {Text Classification from Labeled and Unlabeled Documents Using EM}.
\newblock {\em Machine learning}, 39:103--134.

\bibitem[\protect\citename{Paul and Dredze}2015]{TACL403}
Michael Paul and Mark Dredze.
\newblock 2015.
\newblock {SPRITE: Generalizing Topic Models with Structured Priors}.
\newblock {\em Transactions of the Association for Computational Linguistics},
  3:43--57.

\bibitem[\protect\citename{Pennington \bgroup et al.\egroup
  }2014]{Pennington14}
Jeffrey Pennington, Richard Socher, and Christopher Manning.
\newblock 2014.
\newblock {Glove: Global Vectors for Word Representation}.
\newblock In {\em Proceedings of the 2014 Conference on Empirical Methods in
  Natural Language Processing}, pages 1532--1543.

\bibitem[\protect\citename{Petterson \bgroup et al.\egroup
  }2010]{NIPS2010_4094}
James Petterson, Wray Buntine, Shravan~M. Narayanamurthy, Tib\'{e}rio~S.
  Caetano, and Alex~J. Smola.
\newblock 2010.
\newblock {Word Features for Latent Dirichlet Allocation}.
\newblock In {\em Advances in Neural Information Processing Systems 23}, pages
  1921--1929.

\bibitem[\protect\citename{Phan \bgroup et al.\egroup }2011]{Phan2011HTF}
Xuan-Hieu Phan, Cam-Tu Nguyen, Dieu-Thu Le, Le-Minh Nguyen, Susumu Horiguchi,
  and Quang-Thuy Ha.
\newblock 2011.
\newblock {A Hidden Topic-Based Framework Toward Building Applications with
  Short Web Documents}.
\newblock {\em IEEE Transactions on Knowledge and Data Engineering},
  23(7):961--976.

\bibitem[\protect\citename{Platt}1999]{Platt:1999}
John~C. Platt.
\newblock 1999.
\newblock {Fast Training of Support Vector Machines using Sequential Minimal
  Optimization}.
\newblock In Bernhard Sch\"{o}lkopf, Christopher J.~C. Burges, and Alexander~J.
  Smola, editors, {\em Advances in kernel methods}, pages 185--208.

\bibitem[\protect\citename{Porteous \bgroup et al.\egroup
  }2008]{PorteousNIASW2008}
Ian Porteous, David Newman, Alexander Ihler, Arthur Asuncion, Padhraic Smyth,
  and Max Welling.
\newblock 2008.
\newblock {Fast Collapsed Gibbs Sampling for Latent Dirichlet Allocation}.
\newblock In {\em Proceedings of the 14th ACM SIGKDD international conference
  on Knowledge discovery and data mining}, pages 569--577.

\bibitem[\protect\citename{Robert and Casella}2004]{Robert2004}
Christian~P. Robert and George Casella.
\newblock 2004.
\newblock {\em {Monte Carlo Statistical Methods (Springer Texts in
  Statistics)}}.
\newblock Springer-Verlag New York, Inc.

\bibitem[\protect\citename{Sahami and Heilman}2006]{Sahami20006}
Mehran Sahami and Timothy~D. Heilman.
\newblock 2006.
\newblock {A Web-based Kernel Function for Measuring the Similarity of Short
  Text Snippets}.
\newblock In {\em Proceedings of the 15th International Conference on World
  Wide Web}, pages 377--386.

\bibitem[\protect\citename{Salakhutdinov and Hinton}2009]{NIPS2009_3856}
Ruslan Salakhutdinov and Geoffrey Hinton.
\newblock 2009.
\newblock {Replicated Softmax: an Undirected Topic Model}.
\newblock In {\em Advances in Neural Information Processing Systems 22}, pages
  1607--1614.

\bibitem[\protect\citename{Socher \bgroup et al.\egroup }2013]{socherEtAl2013}
Richard Socher, John Bauer, Christopher~D. Manning, and Ng~Andrew~Y.
\newblock 2013.
\newblock {Parsing with Compositional Vector Grammars}.
\newblock In {\em Proceedings of the 51st Annual Meeting of the Association for
  Computational Linguistics (Volume 1: Long Papers)}, pages 455--465.

\bibitem[\protect\citename{Srivastava \bgroup et al.\egroup }2013]{Nitish2013}
Nitish Srivastava, Ruslan Salakhutdinov, and Geoffrey Hinton.
\newblock 2013.
\newblock {Modeling Documents with a Deep Boltzmann Machine}.
\newblock In {\em Proceedings of the Twenty-Ninth Conference on Uncertainty in
  Artificial Intelligence}, pages 616--624.

\bibitem[\protect\citename{Stevens \bgroup et al.\egroup }2012]{Stevens2012}
Keith Stevens, Philip Kegelmeyer, David Andrzejewski, and David Buttler.
\newblock 2012.
\newblock {Exploring Topic Coherence over Many Models and Many Topics}.
\newblock In {\em Proceedings of the 2012 Joint Conference on Empirical Methods
  in Natural Language Processing and Computational Natural Language Learning},
  pages 952--961.

\bibitem[\protect\citename{Teh \bgroup et al.\egroup }2006]{TehNW2006}
Yee~W Teh, David Newman, and Max Welling.
\newblock 2006.
\newblock {A Collapsed Variational Bayesian Inference Algorithm for Latent
  Dirichlet Allocation}.
\newblock In {\em Advances in Neural Information Processing Systems 19}, pages
  1353--1360.

\bibitem[\protect\citename{Toutanova and Johnson}2008]{NIPS2007_964}
Kristina Toutanova and Mark Johnson.
\newblock 2008.
\newblock {A Bayesian LDA-based Model for Semi-Supervised Part-of-speech
  Tagging}.
\newblock In {\em Advances in Neural Information Processing Systems 20}, pages
  1521--1528.

\bibitem[\protect\citename{Vitale \bgroup et al.\egroup }2012]{Vitale2012}
Daniele Vitale, Paolo Ferragina, and Ugo Scaiella.
\newblock 2012.
\newblock {Classification of Short Texts by Deploying Topical Annotations}.
\newblock In {\em Proceedings of the 34th European Conference on Advances in
  Information Retrieval}, pages 376--387.

\bibitem[\protect\citename{Weng \bgroup et al.\egroup }2010]{Weng2010}
Jianshu Weng, Ee-Peng Lim, Jing Jiang, and Qi~He.
\newblock 2010.
\newblock {TwitterRank: Finding Topic-sensitive Influential Twitterers}.
\newblock In {\em Proceedings of the Third ACM International Conference on Web
  Search and Data Mining}, pages 261--270.

\bibitem[\protect\citename{Xie and Xing}2013]{Pengtao13}
Pengtao Xie and Eric~P. Xing.
\newblock 2013.
\newblock {Integrating Document Clustering and Topic Modeling}.
\newblock In {\em Proceedings of the Twenty-Ninth Conference on Uncertainty in
  Artificial Intelligence}, pages 694--703.

\bibitem[\protect\citename{Yan \bgroup et al.\egroup }2013]{Yan2013}
Xiaohui Yan, Jiafeng Guo, Yanyan Lan, and Xueqi Cheng.
\newblock 2013.
\newblock {A Biterm Topic Model for Short Texts}.
\newblock In {\em Proceedings of the 22Nd International Conference on World
  Wide Web}, pages 1445--1456.

\bibitem[\protect\citename{Yin and Wang}2014]{Yin2014}
Jianhua Yin and Jianyong Wang.
\newblock 2014.
\newblock {A Dirichlet Multinomial Mixture Model-based Approach for Short Text
  Clustering}.
\newblock In {\em Proceedings of the 20th ACM SIGKDD International Conference
  on Knowledge Discovery and Data Mining}, pages 233--242.

\bibitem[\protect\citename{Zhai and Boyd-graber}2013]{ICML2013_zhai13}
Ke~Zhai and Jordan~L. Boyd-graber.
\newblock 2013.
\newblock {Online Latent Dirichlet Allocation with Infinite Vocabulary}.
\newblock In {\em Proceedings of the 30th International Conference on Machine
  Learning}, pages 561--569.

\bibitem[\protect\citename{Zhao \bgroup et al.\egroup }2011]{Zhao2011}
Wayne~Xin Zhao, Jing Jiang, Jianshu Weng, Jing He, Ee-Peng Lim, Hongfei Yan,
  and Xiaoming Li.
\newblock 2011.
\newblock {Comparing Twitter and Traditional Media Using Topic Models}.
\newblock In {\em Proceedings of the 33rd European Conference on Advances in
  Information Retrieval}, pages 338--349.

\end{thebibliography}

\end{document}